\documentclass[sigconf, screen]{acmart}

\usepackage{subcaption}
\usepackage{tabu}
\usepackage{threeparttable}
\usepackage{tablefootnote}

\AtBeginDocument{%
  }

\setcopyright{acmlicensed}
\copyrightyear{2025}
\acmYear{2025}
\acmDOI{XXXXXXX.XXXXXXX}
\acmConference[Conference acronym 'XX]{Make sure to enter the correct
  conference title from your rights confirmation email}{June 03--05,
  2018}{Woodstock, NY}
\acmISBN{978-1-4503-XXXX-X/2018/06}

\acmSubmissionID{7151}

\begin{document}

\title{MCBlock: Boosting Neural Radiance Field Training Speed by MCTS-based Dynamic-Resolution Ray Sampling}

\author{Yunpeng Tan}
\authornote{Both authors contributed equally to this research.}
\affiliation{%
  \institution{Peking University}
  \city{Beijing}
  \country{China}}
\email{yunpengtan@stu.pku.edu.cn}

\author{Junlin Hao}
\affiliation{%
  \institution{Peking University}
  \city{Beijing}
  \country{China}}
\email{2401112126@stu.pku.edu.cn}

\author{Jiangkai Wu}
\affiliation{%
  \institution{Peking University}
  \city{Beijing}
  \country{China}}
\email{jiangkai.wu@stu.pku.edu.cn}

\author{Liming Liu}
\affiliation{%
  \institution{Peking University}
  \city{Beijing}
  \country{China}}
\email{llm@stu.pku.edu.cn}

\author{Qingyang Li}
\affiliation{%
  \institution{Peking University}
  \city{Beijing}
  \country{China}}
\email{liqingyang@stu.pku.edu.cn}

\author{Xinggong Zhang}
\authornote{Corresponding author}
\affiliation{%
  \institution{Peking University}
  \city{Beijing}
  \country{China}}
\email{zhangxg@pku.edu.cn}


\begin{abstract}
  Neural Radiance Field (NeRF) is widely known for high-fidelity novel view synthesis. However, even the state-of-the-art NeRF model, Gaussian Splatting, requires minutes for training, far from the real-time performance required by multimedia scenarios like telemedicine. One of the obstacles is its inefficient sampling, which is only partially addressed by existing works. Existing point-sampling algorithms uniformly sample simple-texture regions (easy to fit) and complex-texture regions (hard to fit), while existing ray-sampling algorithms sample these regions all in the finest granularity (i.e. the pixel level), both wasting GPU training resources. Actually, regions with different texture intensities require different sampling granularities. To this end, we propose a novel dynamic-resolution ray-sampling algorithm, MCBlock, which employs Monte Carlo Tree Search (MCTS) to partition each training image into pixel blocks with different sizes for active block-wise training. Specifically, the trees are initialized according to the texture of training images to boost the initialization speed, and an expansion/pruning module dynamically optimizes the block partition. MCBlock is implemented in Nerfstudio, an open-source toolset, and achieves a training acceleration of up to 2.33x, surpassing other ray-sampling algorithms. We believe MCBlock can apply to any cone-tracing NeRF model and contribute to the multimedia community.
\end{abstract}



\received{20 February 2007}
\received[revised]{12 March 2009}
\received[accepted]{5 June 2009}

\maketitle

\section{Introduction}
\label{sec:introduction}

    \begin{figure}
        \begin{subfigure}[t]{0.22\textwidth}
            \centering
            \includegraphics[width=\linewidth]{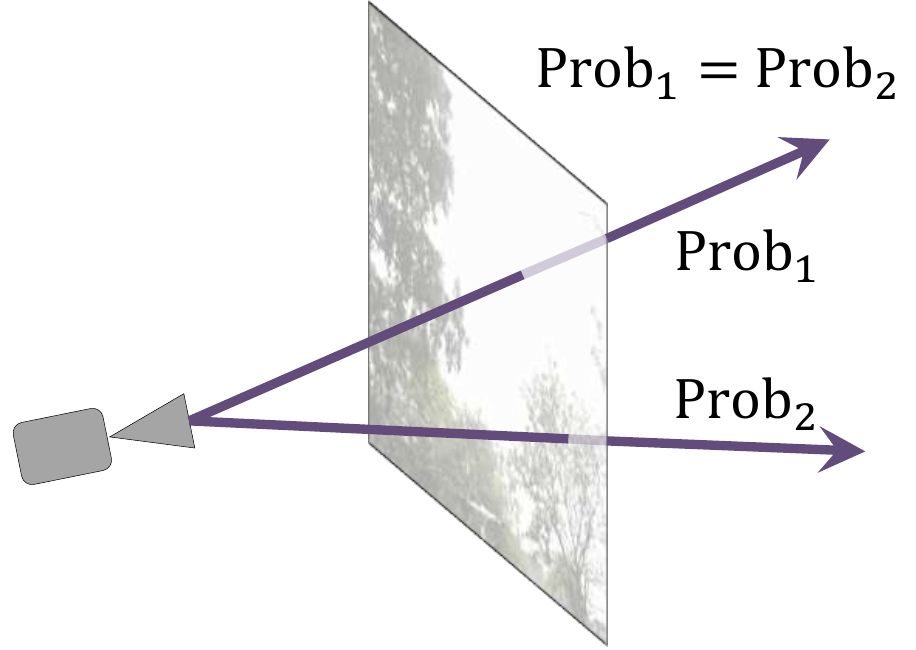}
            \caption{Random Sampling}
            \label{subfig:sampling_random}
        \end{subfigure}
        \hfill
        \begin{subfigure}[t]{0.22\textwidth}
            \centering
            \includegraphics[width=\linewidth]{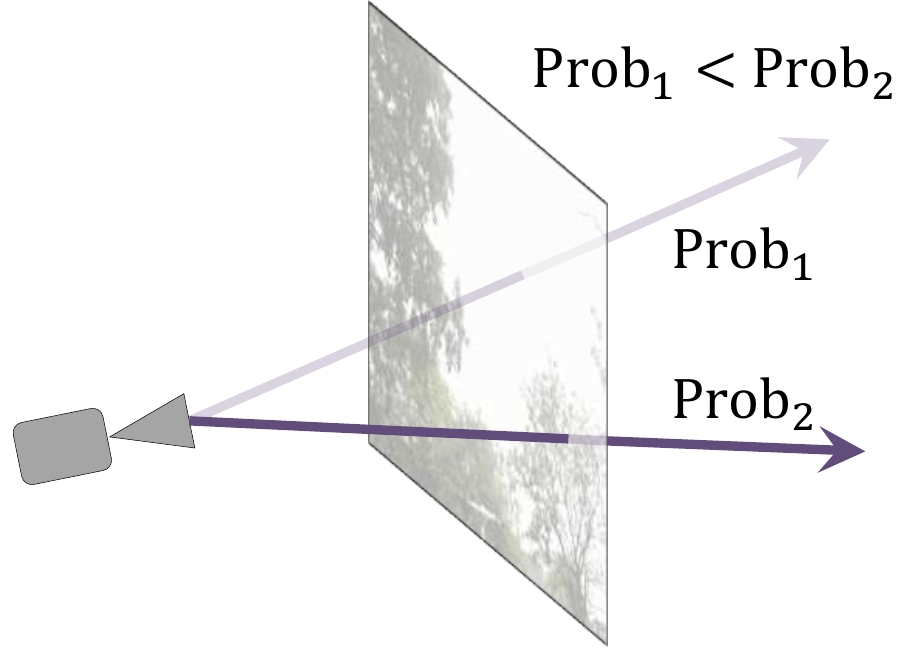}
            \caption{Active Sampling}
            \label{subfig:sampling_actray}
        \end{subfigure}
        \begin{subfigure}[t]{0.22\textwidth}
            \centering
            \includegraphics[width=\linewidth]{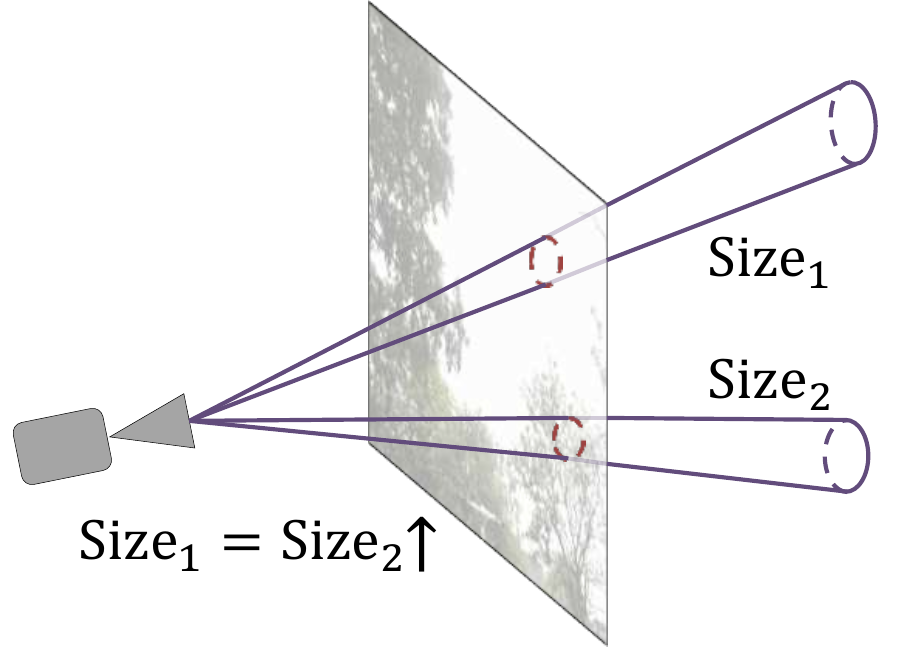}
            \caption{Multi-Resolution Sampling}
            \label{subfig:sampling_C-to-F}
        \end{subfigure}
        \hfill
        \begin{subfigure}[t]{0.22\textwidth}
            \centering
            \includegraphics[width=\linewidth]{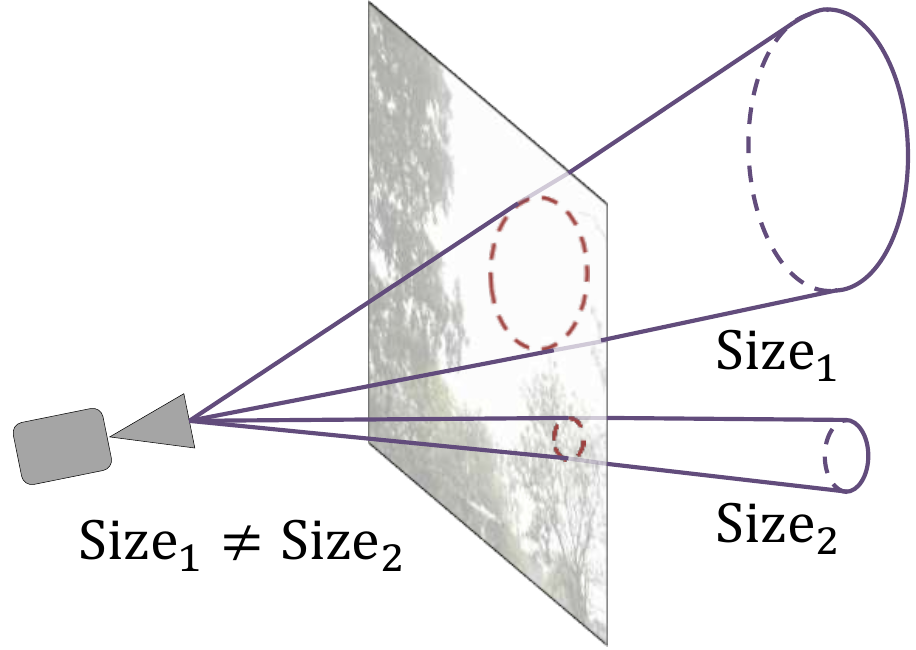}
            \caption{Dynamic-Resolution Sampling}
            \label{subfig:sampling_mcts}
        \end{subfigure}
        \caption{Different ray-sampling algorithms. Random ray sampling \textbf{(a)} samples rays in a uniform random way. Active ray sampling \textbf{(b)} assigns larger sampling probabilities for hard-to-fit pixels. (a)(b) together illustrate the single-resolution ray sampling. The coarse-to-fine multi-resolution ray sampling \textbf{(c)} samples blocks of the same size at the same time, and gradually increases the block size along with training. Dynamic-resolution ray sampling (Ours) \textbf{(d)} samples blocks of different sizes according to the image texture and the training process.}
        \label{fig:diff_method}
    \end{figure}

    3D reconstruction is a core component of multimedia, aiming to transform images or videos into spatial representations, enabling deeper interpretation environments. Its application usually requires real-time performance, such as immersive telepresence, distance education, and metaverse. 
    In recent years, a large number of works combining deep learning and 3D reconstruction have emerged~\cite{muller2021real,323831,1640781}. One of these methods is Neural Radiance Field (NeRF)~\cite{10.1145/3503250}, which employs differentiable volumetric rendering, resulting in novel view synthesis with high fidelity, especially for real-world scenes with complex geometry. However, the slow training speed of NeRF prohibits its practical use in real-time reconstruction.

    One of the main obstacles to real-time NeRF training is its inefficient sampling strategy. The ray marching rendering algorithm employed by NeRF uniformly samples numerous points from the vast 3D space, resulting in extremely low rendering speed. Actually, 3D areas such as air and single-texture objects only require a small amount of sampling, since these areas are easy to fit. However, how to locate the ineffective sampling and reduce it remains a challenge.
    
    Existing sampling algorithms can only partially address the issue of ineffective sampling. Currently, there are two orthogonal types of methods: point sampling and ray sampling. Specifically, \textbf{\textit{point-sampling}} algorithms~\cite{10.1145/3503250,muller2022instant,hu2022efficientnerf} concentrate sampling points near object surfaces, thus learning finer details on object edges. However, these works sample pixels in a uniform random way (Figure~\ref{subfig:sampling_random}). Pixels from simple-texture image regions require less sampling since these regions are always easy to fit~\cite{10.1145/3610548.3618254}. Inspired by this property, ActRay~\cite{10.1145/3610548.3618254} proposes an active \textbf{\textit{ray-sampling}} algorithm~\cite{10.1145/3610548.3618254,korhonen2025efficient,fukuda2024important,sun2024efficient} that performs more frequent sampling in hard-to-fit pixels (Figure~\ref{subfig:sampling_actray}). However, it still needs to constantly sample pixels in simple-texture regions to avoid generating floaters during training. PyNeRF~\cite{turki2024pynerf} applies a coarse-to-fine multi-resolution sampling algorithm. During each training stage, it samples blocks of the same size, and gradually increases the block size along with training. It aims to use the coarse geometry gained in the fast low-resolution training process to guide the point-sampling algorithm in the following high-resolution training process. However, it samples easy-to-fit areas as frequently as hard-to-fit areas, causing redundant sampling in easy-to-fit areas (e.g., simple-texture regions), in the end deteriorating training efficiency.
    
    It’s feasible to further improve ray-sampling efficiency by applying different sampling granularities for regions with different texture intensities. As shown by cone-tracing NeRFs~\cite{Barron_2021_ICCV,Barron_2022_CVPR,Barron_2023_ICCV,isaac2023exact,huang2023local}, training a low-resolution pixel (i.e., a block of high-resolution pixels) means training the low-frequency model parameters (Section~\ref{sec:preliminary}). Besides, simple-texture image regions usually correspond to low-frequency 3D space regions, which can be fit using only low-frequency model parameters. Thus, for simple-texture regions, we only need to sample low-resolution pixels for training. For the complex-texture image regions, we still need to sample high-resolution pixels to learn the high-frequency details.
    
    Motivated by this, we propose dynamic-resolution ray sampling, which partitions training images into pixel blocks with different sizes according to their texture intensities, and trains one simple-texture region as a whole, as shown in Figure~\ref{subfig:sampling_mcts}. 
    To conduct dynamic-resolution sampling, this work has addressed the following technical challenges: First, the initial block size is important for training speed, since sampling a huge block as one cone is useless for 3D scene reconstruction, while initiating training with high-resolution pixels leads to suboptimal efficiency.
    Second, there's a trade-off between block size and training quality. If the block size is too large, it will overlook high-frequency details, leading to a decrease in rendering quality. If the block size is too small, it will conduct useless training on high-frequency parameters, leading to a decrease in training speed.

    To overcome these challenges, we propose MCBlock, a Monte-Carlo-Tree-Search-based~\cite{10.1007/978-3-540-75538-8_7,10.1007/11871842_29} (MCTS) active dynamic-resolution sampling algorithm, where each tree node corresponds to a pixel block.
    In the beginning, we directly initialize a block-tree structure considering the texture intensity of training images. 
    During training, to search for the optimal block partition, Monte-Carlo trees are dynamically expanded and pruned to simulate block splitting and merging. 


    In this paper, we make the following contributions:
    \begin{itemize}
        \item We propose a texture-based tree structure initialization to get appropriately sized initial blocks.
        \item We propose adaptive node expansion/pruning to dynamically search for the optimal block partition along with the training process.
        \item We evaluate MCBlock using different cone-tracing NeRF models and datasets, reaching an acceleration of up to \textbf{2.33x}. We also compare MCBlock with other sampling algorithms, demonstrating its superiority and necessity.
    \end{itemize}




\section{Related Work}
\label{sec:related_work}

\subsection{Neural Radiance Field}
\label{subsec:neural_radiance_field}

    For a scene, Neural Radiance Field (NeRF)~\cite{10.1145/3503250} applies differentiable volumetric rendering to learn a continuous volumetric field which is represented as MLPs. Due to the ability of its MLPs to store high-frequency information, it can render novel view images with higher fidelity, boosting a large number of works like 3D content generation~\cite{Schwarz20neurips_graf,Yu20arxiv_pixelNeRF,na2024uforecon,li2024know}, illumination modeling~\cite{Boss20arxiv_NeRD,Srinivasan20arxiv_NeRV,zhang2021nerfactor,jin2023tensoir,verbin2022ref}, scene editing~\cite{zhang2021stnerf,wang2022clip,bao2023sine,instructnerf2023} and so on. Despite the high rendering quality, the slow training speed prevents NeRF from being applied to scenarios that require fast reconstruction.

\subsection{NeRF Training Acceleration}
\label{subsec:nerf_training_acceleration}

    Currently, the majority of NeRF training acceleration can be categorized into two approaches: 3D representation optimization and sampling optimization. The latter can be further divided into point-sampling and ray-sampling algorithms. Also, there are other acceleration tools~\cite{yong2024gl}.

\subsubsection{3D Representaion Optimization}
\label{subsubsec:3d_representation_optimization}

    Some works propose new NeRF models to accelerate the training process. Plenoxels~\cite{fridovich2022plenoxels}, Instant-NGP~\cite{muller2022instant}, NSVF~\cite{liu2020neural}, DVGO~\cite{SunSC22}, TensoRF~\cite{chen2022tensorf} etc~\cite{kato2024plug,yang2024clear}. use explicit voxel grids or planes to represent a 3D scene, which constrain the gradients in local space, greatly promoting the training speed. However, these methods apply random pixel-wise sampling, ignoring the different influences brought by training different rays.
    
    Gaussian Splatting~\cite{kerbl20233d} leverages a fast differentiable rasterization pipeline to significantly accelerate both the training and rendering processes. During Gaussian sorting, it evenly divides each image into large pixel blocks for parallelization but overlooks variations in texture intensity, resulting in suboptimal efficiency. Specifically, for blocks of the same size, those with more complex textures contain more Gaussians, which decreases the overall sorting speed. The dynamic resolution of MCBlock can be applied here to further accelerate the training and rendering of Gaussian Splatting.
    
\subsubsection{Point-Sampling Algorithms}
\label{subsubsec:point_sampling_algorithm}
    
    Point-sampling algorithms~\cite{10.1145/3503250,muller2022instant,hu2022efficientnerf,wu2024hi,li2024l0,bello2024pronerf,yoo2024improving} estimate the location of the surface of each ray, and sample points near the surfaces, which can focus training on the surface details instead of well-trained air space. However, they still select training rays in a uniform random way. This leads to oversampling of easy-to-fit rays and undersampling of harder-to-fit rays, decreasing the training efficiency.

\subsubsection{Ray-Sampling Algorithms}
\label{subsubsec:ray_sampling_algorithms}

    Ray-sampling algorithms operate in image space. To form a training batch, they either focus on selecting which rays to sample (probabilistic ray sampling), or on adjusting the granularity of ray sampling (multi-resolution ray sampling).

    
    \textbf{Probabilistic Ray Sampling:} Most NeRF works sample rays in a uniform \textbf{\textit{random}} way, needlessly conducting massive sampling on easy-to-fit rays. ActRay~\cite{10.1145/3610548.3618254} proposes an \textit{\textbf{active}} sampling algorithm that gives priority to hard-to-fit rays. Specifically, it uses its redesigned Upper Confidence Bound (UCB)~\cite{slivkins2019introduction} formula for each ray as the sampling probability. Through this ray-sampling algorithm, ActRay can reconstruct scenes with finer details and illumination effects in the early training stage. Similarly, Soft Mining~\cite{kheradmand2024accelerating} uses Langevin Monte-Carlo~\cite{brosse2018promises,neal2011mcmc}(a Stochastic Process algorithm) to select training rays for faster reconstruction. But they still train the NeRF model in a pixel-wise way, unnecessarily learning high-frequency parameters for simple-texture image regions. Motivated by them, we also apply active sampling in MCBlock.


    \textbf{Multi-Resolution Ray Sampling}: The majority of NeRF works sample rays at a \textbf{\textit{single resolution}} for training, training high-frequency information for all image regions without regard to their texture intensity. This unnecessarily trains the high-frequency parameters for simple-texture image regions. Since a low-resolution pixel can be simply represented as a block of high-resolution pixels, some works employ \textbf{\textit{multi-resolution}} sampling~\cite{Barron_2021_ICCV,Srinivasan20arxiv_NeRV,turki2024pynerf}. Among them, PyNeRF~\cite{turki2024pynerf} uses the image coarse-to-fine training, using the coarse geometry gained in the fast low-resolution training process to guide the point-sampling algorithm in the following high-resolution training process. But it equally treats every image region and ignores their difference in texture intensity and fitting difficulty. Motivated by this insight, we propose the \textbf{\textit{dynamic-resolution}} sampling algorithm, MCBlock.

\section{Preliminary}
\label{sec:preliminary}



\label{para:cone_tracing}

    \textbf{Cone-Tracing NeRF}: Actually, each pixel renders one cone region in space, but the original NeRF idealizes one pixel as one point and casts a ray for each pixel, which leads to "jaggies" when rendering low-resolution images. Some works \cite{Barron_2021_ICCV,Barron_2022_CVPR,Barron_2023_ICCV,isaac2023exact,huang2023local} use cone tracing to solve this problem. For example, Mipnerf \cite{Barron_2021_ICCV} casts a cone for each pixel, and splits the cone into conical frustums. Its MLPs output the colors and densities of frustums instead of points. For each conical frustum, Mipnerf encodes it using Integrated Positional Encoding (IPE) into a vector of $\sin$s and $\cos$s, and each term of this vector has an additional coefficient. When the pixel area grows, the coefficients of the last terms will approach zero. In this way, a large frustum will be encoded into only low-frequency space coordinates, thus training only low-frequency information of MLPs, but having no influence on the high-frequency information. With this property, we cast cones that cover our blocks, training parameters with different frequencies according to different block sizes.

\section{Method}
\label{sec:method}

\subsection{MCBlock Training Framework}
\label{subsec:mcblock_training_framework}

    \begin{figure*}[t]
      \centering
      \includegraphics[width=1.0\textwidth]{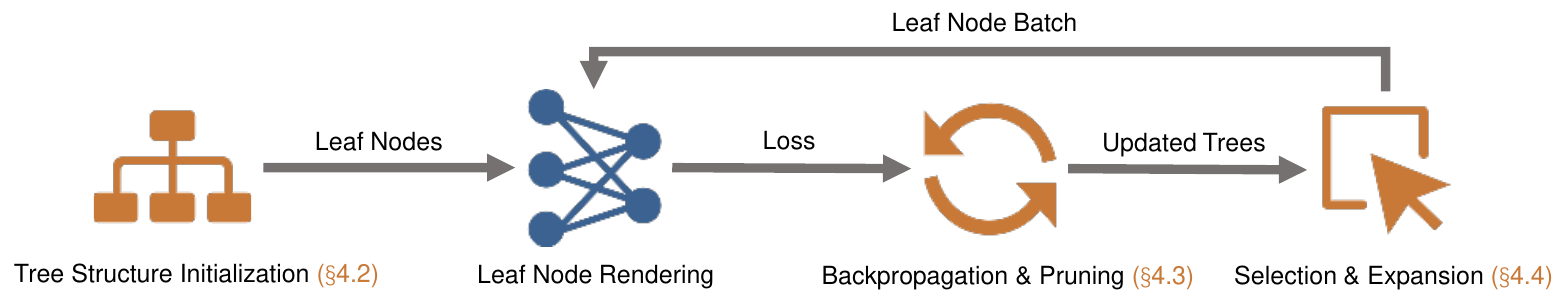}
      \caption{\textbf{Training pipeline of MCBlock.} We first initialize the tree structures using the training images, and render all the leaf nodes to get their loss values. From each leaf node to its root node, these loss values are backpropagated, and at the same time we prune some leaf nodes. With the updated trees, we select some leaf nodes, expand them, and choose a subset of the expanded nodes as a training batch. "Selection \& expansion", "Leaf node rendering", and "Backpropagation \& pruning" will be repeated until the model converges.}
      \label{fig:framework}
    \end{figure*}

    As introduced above, most NeRF works ignore the fact that different image regions need different sampling granularity. For one simple-texture region, we only need to use the entire pixel block for training to reconstruct its corresponding low-frequency 3D space. However, for the complex-texture region, training of small blocks is required to recover the high-frequency details in 3D sprce.

    Inspired by this insight, we propose our MCTS-based dynamic-resolution sampling algorithm. Specifically, each training image corresponds to a Monte-Carlo quadtree, with each node standing for a rectangular pixel block, and its four child nodes standing for four evenly divided sub-blocks.

    Figure~\ref{fig:framework} illustrates our training pipeline. At the beginning of training, we initialize the tree structures using color variances of training images(Section~\ref{subsec:initialization}). Then all leaf nodes are used to train the cone-tracing NeRF model, including loss calculation and gradient descent (Section~\ref{para:cone_tracing}). Next, we backpropagate the rendering losses of these nodes along the path from each leaf node to its corresponding root node and prune some nodes to promote training efficiency (Section~\ref{subsec:backpropagation}). Then, with the updated Monte-Carlo trees, we select leaf nodes using UCT as the probability, and expand the selected leaf nodes. A subset of the expanded nodes is used for training (Section~\ref{subsec:selection}). "Selection \& Expansion", "Leaf node rendering" and "Backpropagation \& Pruning" will be repeated for many iterations, until the model converges.

\subsection{Tree Structure Initialization}
\label{subsec:initialization}

    \begin{figure}
        \centering
        \includegraphics[width=0.45\textwidth]{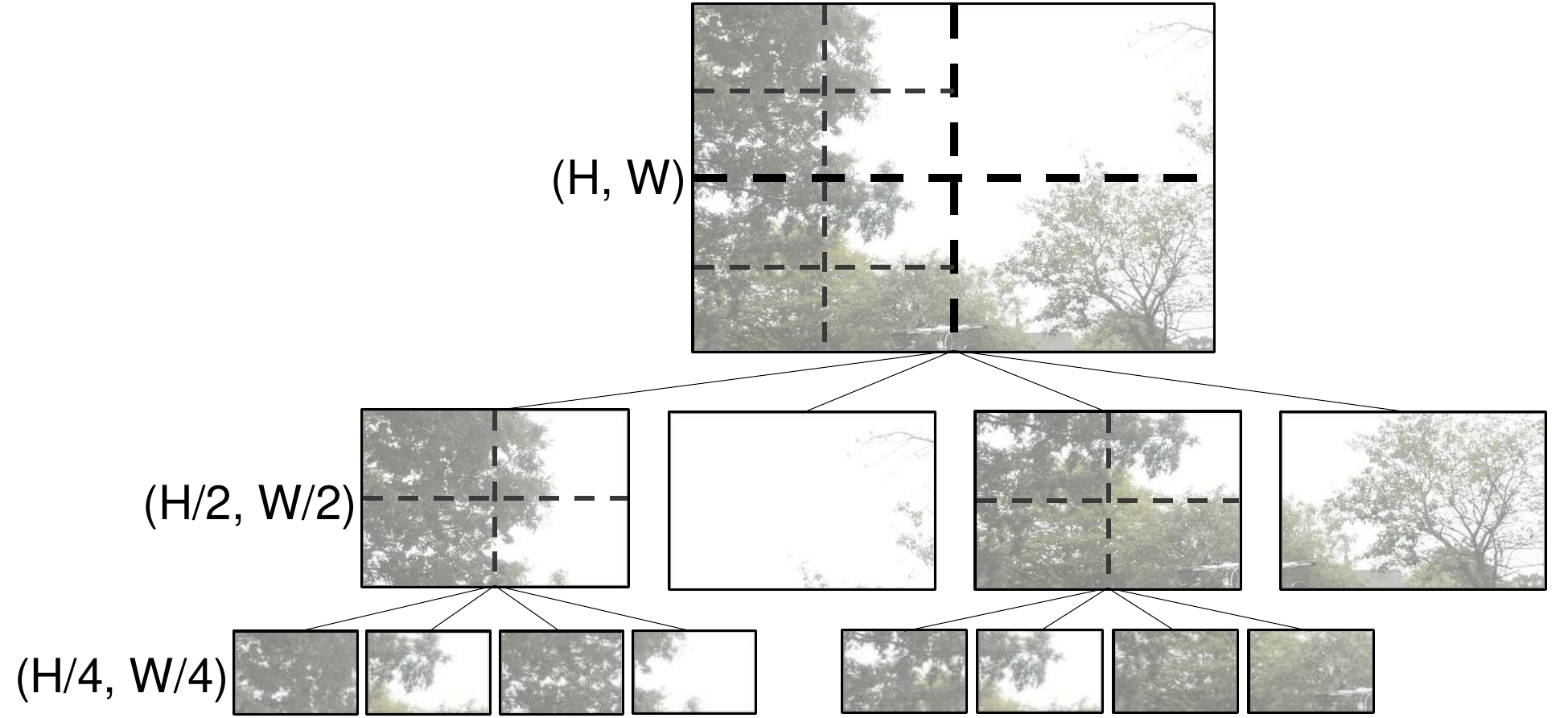}
        \caption{\textbf{Tree structure initialization.} We initialize the structures of trees according to the texture intensity of training images. Regions like the sky have fewer and larger blocks, and regions like the trees have more and smaller blocks.}
        \label{fig:initialization}
    \end{figure}

    To skip the time-consuming initialization process of the traditional MCTS which starts from only root nodes, we directly initialize the tree structures based on the color of training images. For each training image, we first treat all the pixels of the image as the leaves of the tree. Then, we repeatedly merge neighbor sibling nodes. Specifically, for four neighbor sibling nodes, we first add their father node $j$ into the tree. Then, if they are currently leaf nodes and $\mathrm{Variance}(\{C_{p}\}_{p\in B_j}) < \epsilon_{init}$ (where $B_j$ is the block corresponding to node $j$, $C_{p}$ is the color of pixel $p$, and $\epsilon_{init}$ is a constant threshold), they are deleted. The initialization finishes after the root node is added. A toy example of the initialized tree structure is shown in Figure~\ref{fig:initialization}. If the image size is (H, W), the root node stands for the whole image, its four child nodes stand for four evenly divided sub-blocks with the size of (H/2, W/2), its grandchild nodes stand for sub-sub-blocks with the size of (H/4, W/4), and so on. The leaf nodes form an initial block partition of the image.

    After initializing the tree structure, all leaf nodes are fed into a cone-tracing NeRF model in batches. During this process, we not only update the model parameters but also obtain the initial loss values of each node. Although we randomly select subsets of leaf nodes as training batches, this approach still introduces a sampling bias towards regions with complex textures, as these regions are divided into more blocks. Thus training these leaf nodes still poses higher training efficiency than random pixel-wise sampling.

\subsection{Backpropagation \& Pruning}
\label{subsec:backpropagation}

\subsubsection{UCT Redefinition}
\label{subsubsec:redefinition}

    First of all, we need to redesign the UCT formula. That's because the UCT formula of the traditional MCTS~\cite{10.1007/978-3-540-75538-8_7,10.1007/11871842_29} can only be applied to scenarios with static reward distribution, and the UCB formula of ActRay is only designed for pixels instead of blocks, so both of them aren't suitable for our scenario. To fit the block-wise NeRF training scenario, we have redesigned the UCT formulas for the leaf nodes and internal nodes separately. We design the UCT value $U_{i}$ and the loss value $L_{i}$ of \textbf{\textit{leaf node}} $i$ as:
    
    \begin{equation}
    U_{i} = L_{i} * e^{\frac{{O}_{i}}{\lambda}} * \vert B_i \vert,\ L_{i} = \Vert \overline{C_{B_i}} - \widehat{C_{B_i}} \Vert_2
    \label{eq:UCT_leaf}
    \end{equation}
    
    Node $i$ corresponds to block $B_i$. The block rendering loss $L_i$ is defined as the MSE between the average color $\overline{C_{B_i}}$ of block $B_i$ and its most recently rendered color $\widehat{C_{B_i}}$. And it is multiplied by the expired time term $e^{\frac{O_{i}}{\lambda}}$ just like ActRay, where $O_{i}$ is the time interval since node $i$ was last sampled, and $\lambda$ is a temperature coefficient to control the influence of the expired time term. Considering the area of blocks, we add another multiplier $\vert B_i \vert$, i.e. the size (height multiplying width) of block $B_i$.
    
    For an \textbf{\textit{internal node}} $j$, its UCT value $U_{j}$ is defined as the summation of the UCT values of its child nodes $Ch_{j}$, and its loss value $L_{j}$ is defined as the average of the loss values of its child nodes:

    \begin{equation}
    U_{j} = \sum_{k \in Ch_j}U_{k},\ L_{j} = \frac{\sum_{k \in Ch_j}L_{k}}{\vert Ch_{j} \vert}
    \label{eq:UCT_internal}
    \end{equation}


\subsubsection{Backpropagation}
\label{subsubsec:backpropagation}

    \begin{figure}
        \centering
        \includegraphics[width=0.45\textwidth]{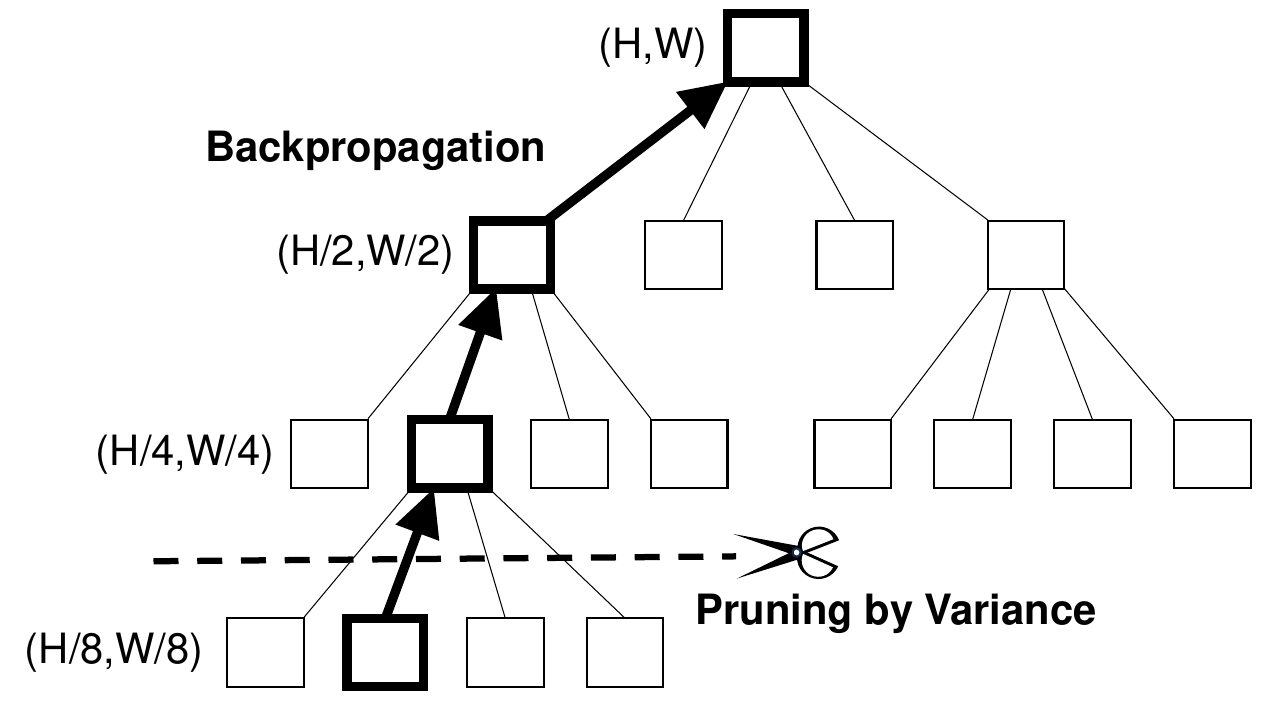}
        \caption{\textbf{Backpropagation \& pruning.} On the path from each leaf node of the batch to its root node, we backpropagate the UCT values and the loss values to nodes on the path and decide whether to prune their child nodes.}
        \label{fig:backpropagation}
    \end{figure}

    Considering efficiency, we don't calculate UCT values on the spot. Instead, we dynamically update them using backpropagation, as shown in Figure~/ref{fig:backpropagation}. Given the loss values of a batch of leaf blocks, we aim to update the UCT value and the loss value of every node. First, we multiply all UCT values by $e^{\frac{1}{\lambda}}$ to simulate the increment of expired time $O$. Then, for each leaf node that corresponds to a block of the batch, we set its UCT value to $L*\vert B \vert$ (its expired time now becomes $0$). Next, we walk up from each leaf node to its root node, and reset each encountered internal node's UCT value and loss value according to the formula in Section~\ref{subsubsec:redefinition}. If batch size is $S_{batch}$ and image size is $S_{image}$, this process only contains complex operations for $\log_4(S_{image})*S_{batch}$ nodes, and simple multiplications for all nodes, ensuring our high efficiency of UCT value maintenance.

\subsubsection{Pruning}
\label{subsubsec:pruning}

    Blocks may be split to eliminate high-frequency flaws (which will be discussed in Section~\ref{subsubsec:expansion}), but after the elimination, some blocks need to be merged again to improve sampling efficiency. We use node pruning to simulate block merging. During backpropagation, for each node on the walking path, we also decide whether to prune its four child nodes, as shown in Figure~\ref{fig:backpropagation}. In detail, for a node $j$, after the update of its UCT value and loss value, we prune its four child nodes $Ch_j$ when $Ch_j$ are all leaf nodes and:
    
    \begin{equation}
    \frac{L_j}{\overline{L}} < \epsilon_{L}\ \ and\ \ \mathrm{Variance}(\{C_{p}\}_{p\in B_j}) < \epsilon_{C}
    \label{eq:merge_condition}
    \end{equation}
    
    Where $\epsilon_{L}$ and $\epsilon_{C}$ are thresholds respectively for the scaled loss and the color variance, and $\overline{L}=\frac{\sum_{i \in \mathrm{NODES}}L_i}{\vert \mathrm{NODES} \vert}$ is the mean loss value of all blocks, designed to eliminate loss scale differences between scenes. Compared with block partition only based on color variance, loss values change along with training, so they can perceive when the high-frequency flaws emerge or vanish, and then split blocks to eliminate flaws or merge blocks to promote efficiency.
    

    Backpropagation updates the values of the tree nodes, and pruning updates the structure of the Monte-Carlo trees.

\subsection{Selection \& Expansion}
\label{subsec:selection}

\subsubsection{Selection}
\label{subsubsec:selection}

    \begin{figure}
        \centering
        \includegraphics[width=0.45\textwidth]{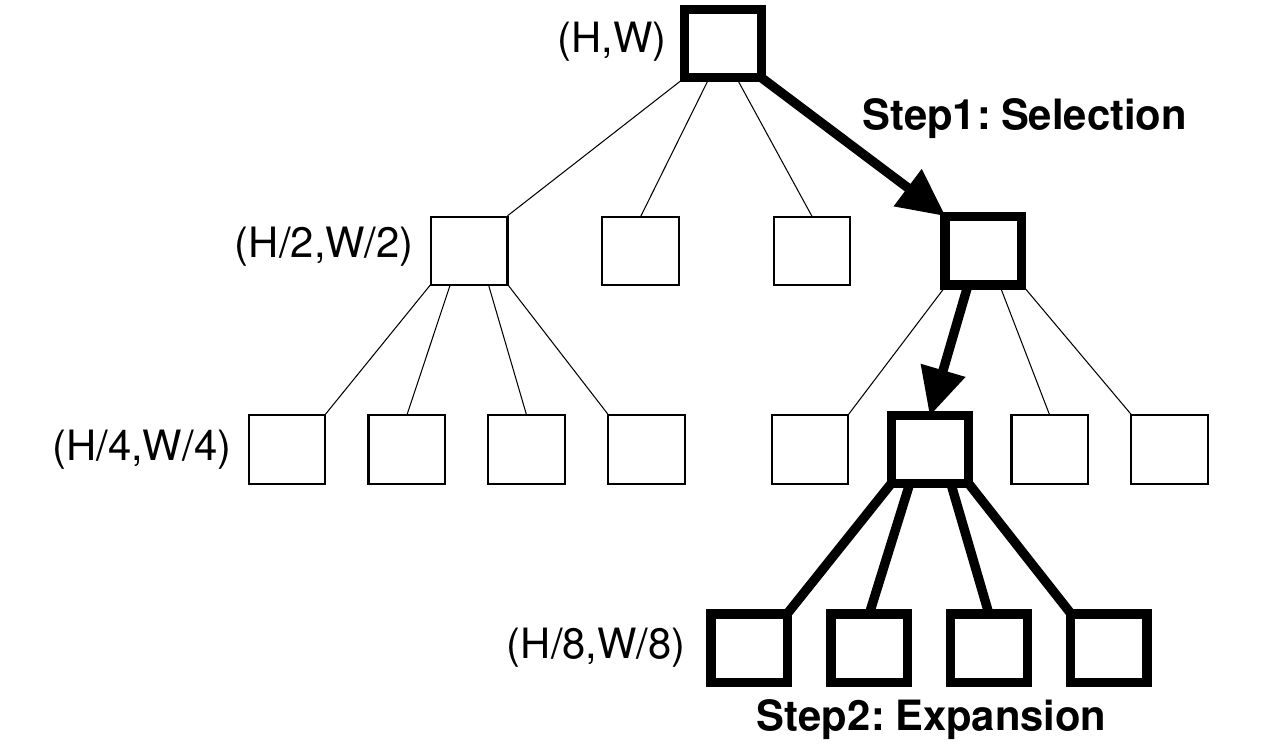}
        \caption{\textbf{Selection \& expansion.} Using UCT as probability, we choose a root node and walk from the root node to one leaf node. Then we expand this node and randomly add one of the expanded nodes into the training batch.}
        \label{fig:selection}
    \end{figure}

    Given the updated trees, to select hard-to-fit blocks to accelerate training, we design a parallel selection algorithm. We select blocks under the guidance of UCT as shown in Figure~\ref{fig:selection}. To select one leaf block, first, we choose one of the root nodes. The probability of choosing root $r$ is $\frac{U_r}{\sum_{s \in \mathrm{ROOTS}}U_s}$ where $\mathrm{ROOTS}$ is the set of all root nodes. Then we walk from the chosen root node to a particular leaf node. At a certain node $j$, for one of its child nodes, node $k$, the probability of stepping to it is set to $\frac{U_k}{U_j}$. This probability sampling process enables GPU parallelization of MCTS, without any additional information to be maintained like A. Liu et~\cite{Liu2020Watch}.

    \begin{figure*}[t]
        \centering
            \begin{subfigure}{0.3\textwidth}
                \centering
                \includegraphics[width=\linewidth]{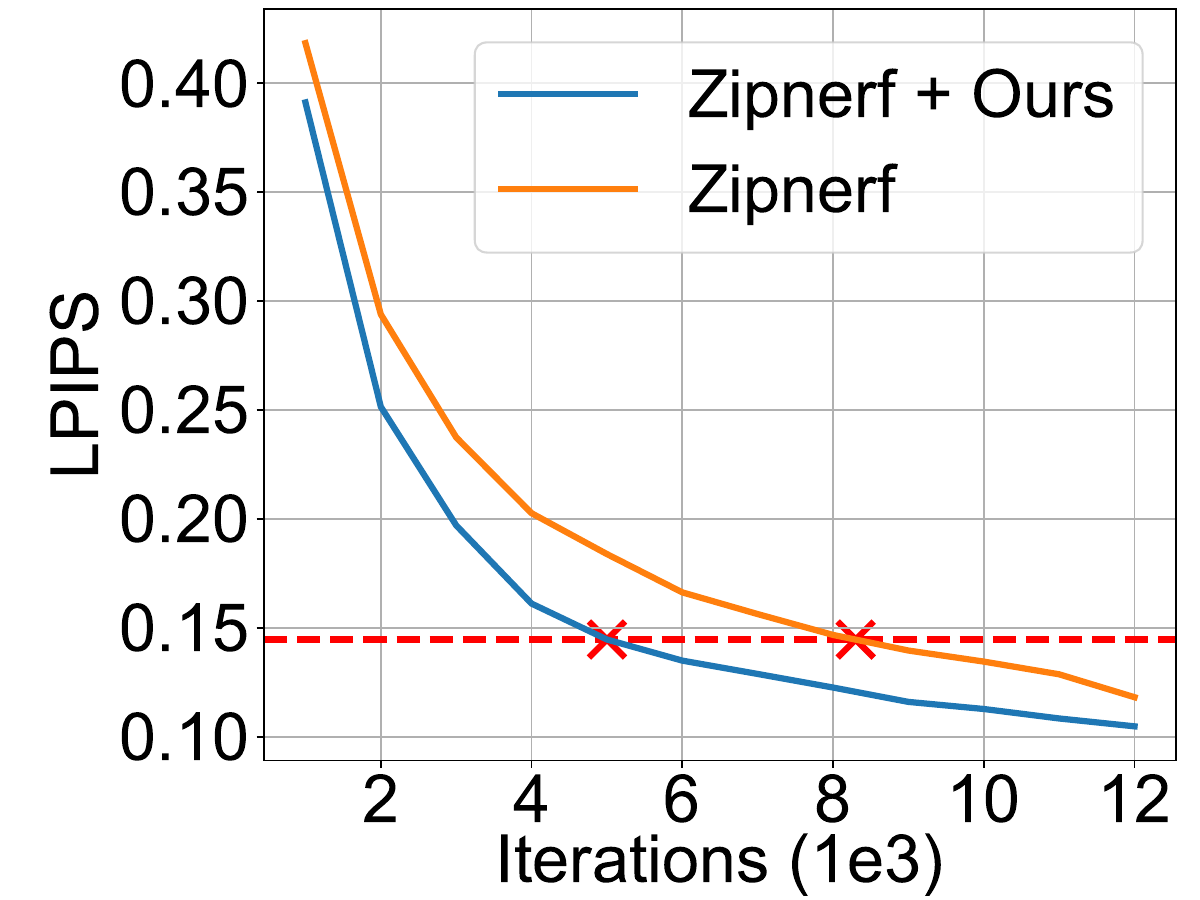}
                \caption{Zipnerf on Mip360 Outdoor}
                \label{subfig:outdoor_lpips}
            \end{subfigure}
            \hfill
            \begin{subfigure}{0.3\textwidth}
                \centering
                \includegraphics[width=\linewidth]{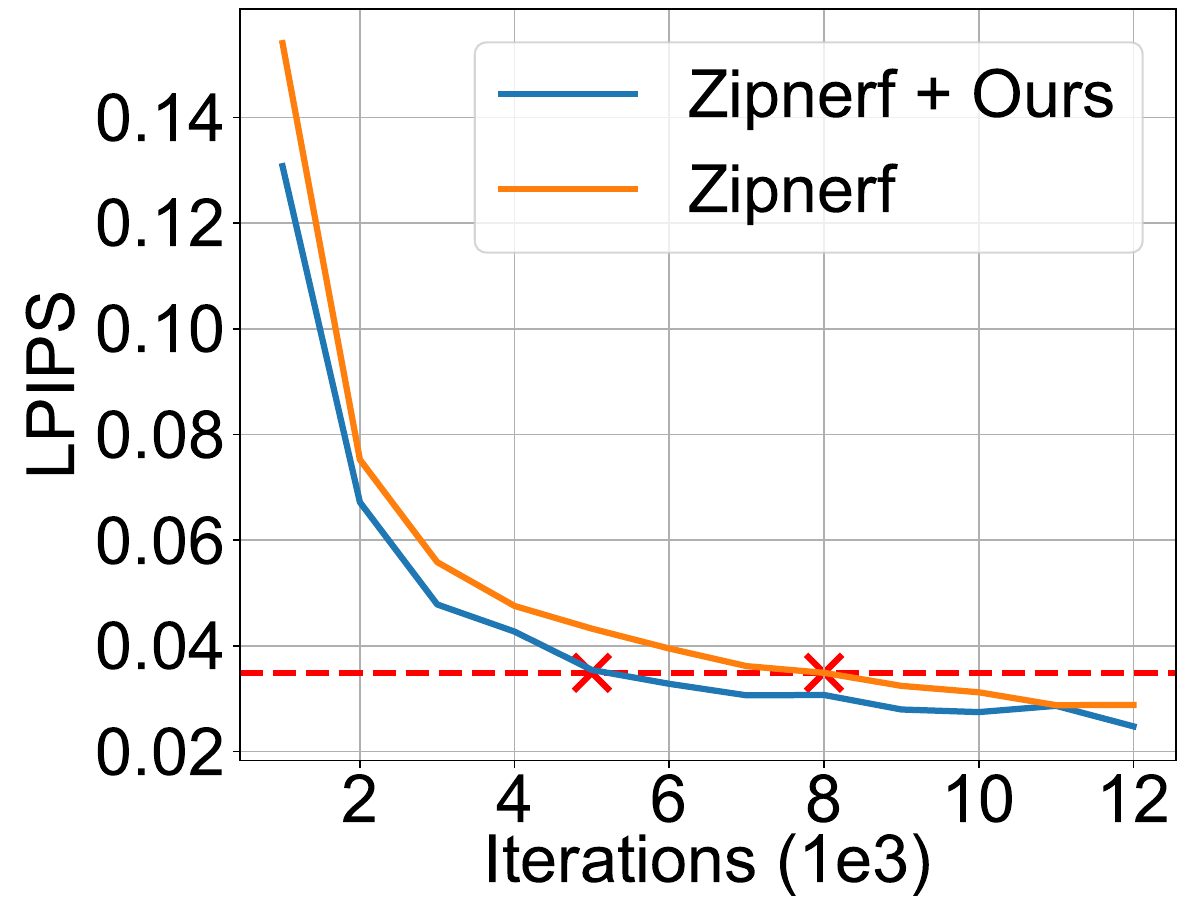}
                \caption{Zipnerf on Mip360 Indoor}
                \label{subfig:indoor_lpips}
            \end{subfigure}
            \hfill
            \begin{subfigure}{0.3\textwidth}
                \centering
                \includegraphics[width=\linewidth]{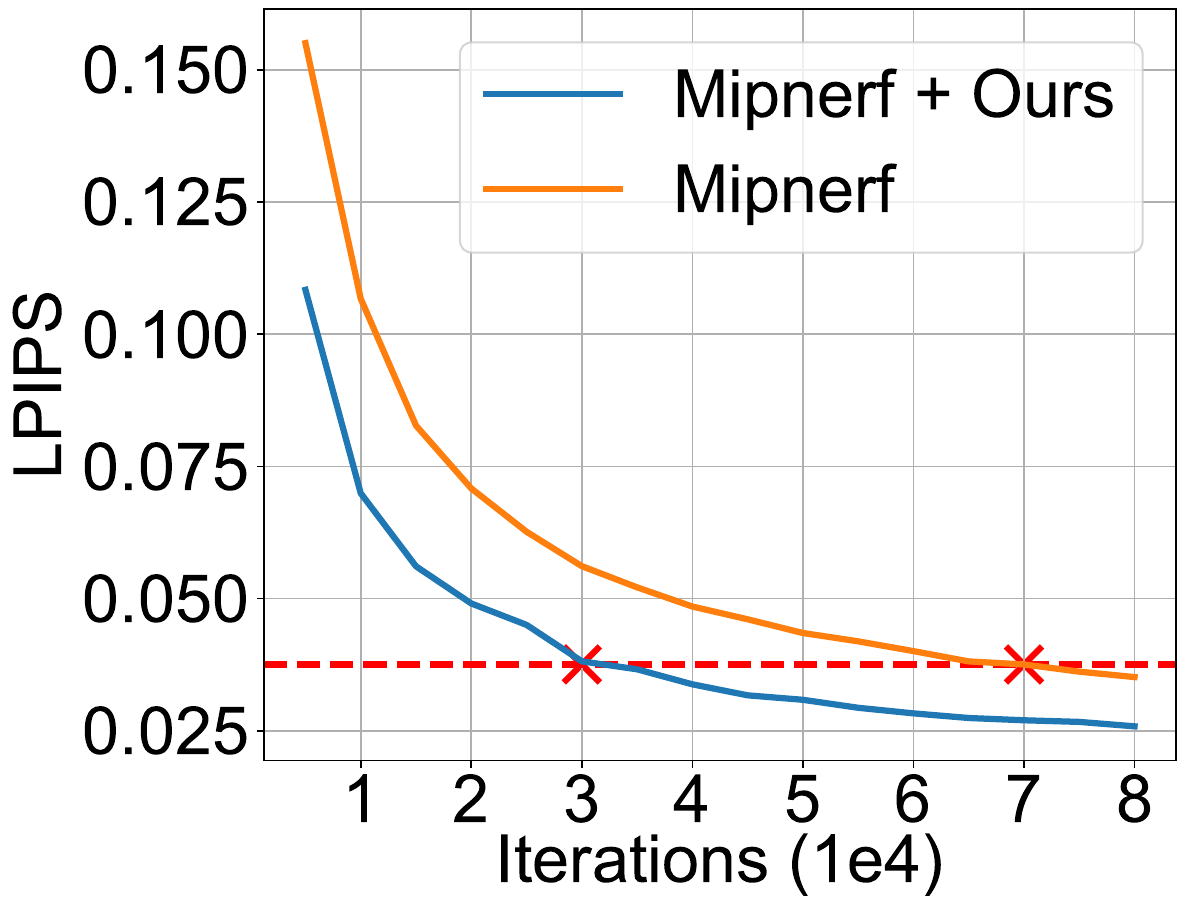}
                \caption{Mipnerf on Blender}
                \label{subfig:blender_lpips}
            \end{subfigure}
        \caption{LPIPS results under different datasets and different NeRF models. With the iteration number as the horizontal axis, our method can accelerate the training process by up to \textbf{2.33x}.}
        \label{fig:diff_fig}
    \end{figure*}

\subsubsection{Expansion}
\label{subsubsec:expansion}

    Due to the lack of supervision of high-frequency parameters in simple-texture regions, high-frequency flaws are likely to emerge. We can eliminate these flaws by splitting large blocks to supervise high-frequency parameters. We use node expansion to simulate block splitting, as shown in Figure~\ref{fig:selection}. Specifically, if a leaf node $j$ is among the selected leaf nodes, it is likely to have a high rendering loss. So we split its corresponding block evenly into four sub-blocks as its four child nodes, trying to detect and finally eliminate the high-frequency flaws. To bypass the time-consuming calculation of accurate losses of the new nodes, we directly assign the UCT value (scaled by the block size) and the loss value of node $j$ to these new nodes. Then we randomly add one of the expanded nodes into the training batch for the training iteration.

    Selection and expansion together generate batches of pixel blocks for training, and update the tree structures.

\section{Experiments}
\label{sec:experiments}

\subsection{Implementation Details}
\label{subsec:imple_detail}

    MCBlock is implemented based on Nerfstudio~\cite{nerfstudio}, an open-source toolset for NeRF. We use the Mipnerf360 dataset~\cite{Barron_2022_CVPR} and the Blender dataset~\cite{10.1145/3503250}, as representatives of real-world datasets and synthetic datasets. Besides, variance thresholds $\epsilon_{init}$, $\epsilon_{L}$ and $\epsilon_{C}$ is respectively set to 1e-3, 1e-2 and 1e-4. We use LPIPS as our primary metric, because LPIPS can better reflect visual quality than PSNR and SSIM. For example, floaters and blurs may not influence PSNR and SSIM, which can influence LPIPS a lot.

    Unlike traditional NeRFs which synthesize novel view images in a \textbf{pixel-wise} way, we render novel view images in a \textbf{block-wise} way. We warp the block partition of the training views to the novel view, and with this estimated block partition of the novel view, we render the novel view image by rendering blocks instead of pixels. This rendering manner can avoid rendering high-frequency flaws that aren't fully eliminated during training, thus reducing artifacts like floaters in the weak-texture regions.


\subsection{Different Datasets}
\label{subsec:diff_models_datasets}

    For each scene of the Mipnerf360 dataset, $\frac{7}{8}$ of images are used for training, and the rest $\frac{1}{8}$ are used for testing. Here we use Zipnerf~\cite{Barron_2023_ICCV} as the NeRF model.
    
    As for the outdoor scenes in the Mipnerf360 dataset, the training images contain many large simple-texture regions such as the sky, which requires fewer sampling times for MCBlock to fit. As illustrated in Figure~\ref{subfig:outdoor_lpips}, compared with random sampling, MCBlock achieves an acceleration of up to \textbf{1.66x} (at the LPIPS = 0.145 line).
    
    As for the indoor scenes in the Mipnerf360 dataset, there are fewer large simple-texture regions, but MCBlock can still find a lot of small simple-texture regions and sample them integrally, reaching an acceleration of up to \textbf{1.6x} (at the LPIPS = 0.035 line) (Figure~\ref{subfig:indoor_lpips}).
    
    For each scene of the Blender dataset, we use 100 images for training and 100 images for testing. Here we use Mipnerf~\cite{Barron_2021_ICCV} as the NeRF model. Due to large background regions, we reach an acceleration of up to 2.33x (at the LPIPS = 0.030 line) (Figure~\ref{subfig:blender_lpips}).

    Unfortunately, the implementation in the Nerfstudio library does not support training Zipnerf on the Blender dataset or training Mipnerf on the Mipnerf360 dataset, so we didn't test these model-dataset combinations.

\subsection{Different Sampling Algorithms}
\label{subsec:diff_algorithm}

    \begin{table*}
    \caption{The performance of different sampling algorithms, using the Blender dataset as dataset and Mipnerf as NeRF model. By applying the point-sampling algorithm of Mipnerf and using MCBlock as the ray-sampling algorithm, we get the best PSNR, SSIM, and LPIPS, especially in the early training stages.}
    \label{tab:diff_algorithm}
    \begin{threeparttable}
    \centering
    \resizebox{\textwidth}{!}
    {
        \begin{tabular} {*{13}{c}}
            \toprule
            Iteration & \multicolumn{3}{c}{20000} & \multicolumn{3}{c}{40000} & \multicolumn{3}{c}{60000} & \multicolumn{3}{c}{80000} \\
            \cmidrule{1-13}
            Method & LPIPS & PSNR & SSIM & LPIPS & PSNR & SSIM & LPIPS & PSNR & SSIM & LPIPS & PSNR & SSIM \\
            \cmidrule(r){1-1}
            \cmidrule(r){2-4}
            \cmidrule(r){5-7}
            \cmidrule(r){8-10}
            \cmidrule{11-13}
            No Strategy & 0.076 & 28.69 & 0.920 & 0.054 & 30.29 & 0.938 & 0.047 & 30.94 & 0.945 & 0.041 & 31.50 & 0.950\\
            Point-Sampling & 0.071 & 29.02 & 0.926 & 0.048 & 30.89 & 0.944 & 0.040 & 31.71 & 0.952 & 0.035 & 32.30 & 0.956 \\
            ActRay & 0.068 & 30.23 & 0.933 & 0.045 & 31.64 & 0.948 & 0.036 & 32.31 & 0.954 & 0.034 & 32.66 & 0.957\\
            Multi-Res\textsuperscript{*} & 0.076 & 28.67 & 0.920 & 0.056 & 30.28 & 0.938 & 0.046 & 30.98 & 0.946 & 0.041 & 31.49 & 0.950\\
            MCBlock & 0.054 & 30.37 & 0.935 & 0.041 & 31.55 & 0.947 & 0.036 & 32.07 & 0.952 & 0.033 & 32.37 & 0.954 \\
            ActRay w/ Point-Sampling & 0.055 & 30.89 & 0.940 & 0.039 & 32.39 & \textbf{0.955} & 0.031 & \textbf{33.20} & \textbf{0.960} & 0.027 & \textbf{33.58} & \textbf{0.963} \\
            Multi-Res w/ Point-Sampling & 0.069 & 29.11 & 0.926 & 0.049 & 30.78 & 0.943 & 0.040 & 31.74 & 0.952 & 0.035 & 32.31 & 0.957 \\
            \midrule
            MCBlock w/ Point-Sampling & \textbf{0.049} & \textbf{31.10} & \textbf{0.942} & \textbf{0.034} & \textbf{32.55} & \textbf{0.955} & \textbf{0.028} & 33.19 & \textbf{0.960} & \textbf{0.026} & 33.52 & 0.962 \\
            
            \bottomrule
        \end{tabular}
    }
    \begin{tablenotes}
    \footnotesize
        \item[*]{Multi-Res stands for multi-resolution ray sampling employed by works like PyNeRF}
    \end{tablenotes}
    \end{threeparttable}
    \end{table*}

    \begin{table*}
    \caption{The performance of different sampling algorithms, using the Mipnerf360 dataset as dataset and Zipnerf as NeRF model.}
    \label{tab:diff_algorithm_mipnerf360}
    \begin{threeparttable}
    \centering
    \resizebox{\textwidth}{!}
    {
        \begin{tabular} {*{19}{c}}
            \toprule
            Iteration & \multicolumn{3}{c}{2000} & \multicolumn{3}{c}{4000} & \multicolumn{3}{c}{6000} & \multicolumn{3}{c}{8000} \\
            \cmidrule{1-13}
            Method & LPIPS & PSNR & SSIM & LPIPS & PSNR & SSIM & LPIPS & PSNR & SSIM & LPIPS & PSNR & SSIM \\
            \cmidrule(r){1-1}
            \cmidrule(r){2-4}
            \cmidrule(r){5-7}
            \cmidrule(r){8-10}
            \cmidrule{11-13}
            Point-Sampling & 0.169 & 27.34 & 0.786 & 0.114 & 29.04 & 0.846 & 0.094 & 29.82 & 0.869 & 0.083 & 30.32 & 0.881\\
            ActRay w/ Point-Sampling & 0.156 & 27.86 & 0.788 & 0.111 & 29.34 & 0.846 & 0.092 & 30.01 & 0.867 & 0.080 & 30.57 & 0.883\\
            Multi-Res w/ Point-Sampling & 0.171 & 27.25 & 0.783 & 0.112 & 29.00 & 0.847 & 0.091 & 29.78 & 0.870 & 0.079 & 30.34 & \textbf{0.884} \\
            \midrule
            MCBlock w/ Point-Sampling & \textbf{0.146} & \textbf{27.96} & \textbf{0.791} & \textbf{0.093} & \textbf{29.45} & \textbf{0.848} & \textbf{0.077} & \textbf{30.27} & \textbf{0.873} & \textbf{0.070} & \textbf{30.66} & \textbf{0.884}\\
            \bottomrule
        \end{tabular}
    }
    \end{threeparttable}
    \end{table*}

    \begin{figure*}[h]
    \centering
        \includegraphics[width=1.0\linewidth]{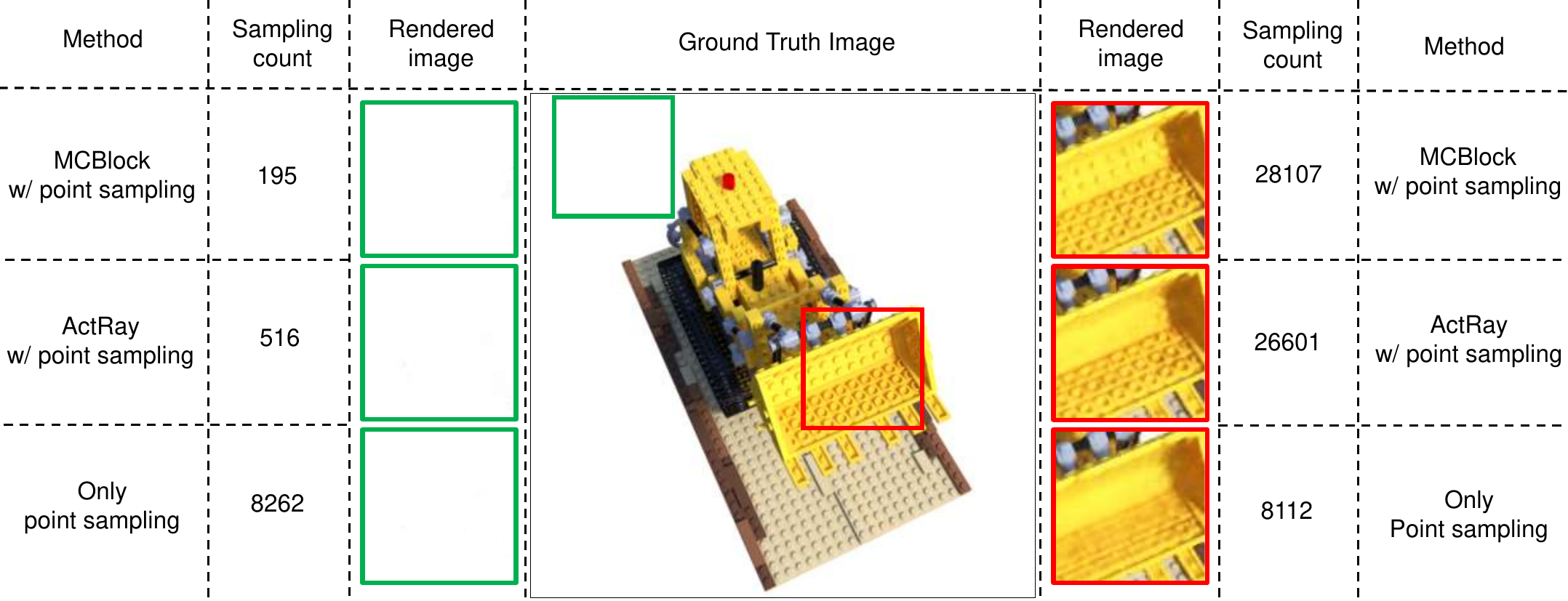}
    \caption{Rendered images and sampling count during 5000 training iterations under the lego scene. On the left, it can be seen that compared with ActRay, MCBlock further decreases the sampling count of the background region, but still renders a clean background without any floater. On the right, it can be seen that MCBlock spares more sampling times for finer foreground details such as clearer bricks of the bucket.}
    \label{fig:mip360_images}
    \end{figure*}

    In this experiment, we use Mipnerf as the NeRF model and the Blender dataset as the dataset. Mipnerf adopts a point-sampling algorithm, which uses proposal sampling to guide the final point sampling and is orthogonal to any ray-sampling algorithm. We apply different ray-sampling algorithms to Mipnerf, respectively the active sampling algorithm ActRay~\cite{10.1145/3610548.3618254}, the multi-resolution sampling algorithm employed by works like PyNeRF~\cite{turki2024pynerf}, and our MCBlock. Table~\ref{tab:diff_algorithm} shows that MCBlock outperforms all other sampling algorithms. Sometimes MCBlock provides slightly lower PSNR and SSIM, because MCBlock might render false highlight effects while other algorithms render no highlight effects. Results of different ray-sampling algorithms on the Mipnerf360 dataset with Zipnerf as the NeRF model are provided in Table~\ref{tab:diff_algorithm_mipnerf360}. Additional subjective results can be found in supplemental material (Section~\ref{sec:subjective_img}).
    
    
    \textbf{Point Sampling (Mipnerf)}: Although sampling points near surfaces can accelerate the training process, applying random ray sampling will conduct too much unnecessary sampling in simple-texture regions, leading to a decrease in training speed. Also, solely applying ray-sampling strategies isn't efficient enough. By combining point sampling and ray sampling (like MCBlock), the training process can reach the highest efficiency.
    
    \label{para:actray}
    
    \textbf{Active Ray Sampling (ActRay)}: ActRay assigns larger sampling probabilities for hard-to-fit pixels to learn better details. We show the rendered images and the cumulative sampling times of MCBlock and ActRay in Figure~\ref{fig:mip360_images}. Although ActRay has decreased the sampling frequency of background regions a lot through active ray sampling, MCBlock further decreased it by training the large blocks of the background in an integrated manner. This spares more sampling times to sample the foreground details, enhancing the foreground's sharpness such as the bricks of the bucket. The advantage of MCBlock over ActRay shrinks in the later iterations, because the loss values are easy to be outdated along with training. The loss propagation strategy of ActRay can help to settle this problem, but it may consume additional time (Section~\ref{subsec:time_conspt}).

    \textbf{Multi-Resolution Ray Sampling (PyNeRF~\cite{turki2024pynerf})}: To validate the multi-resolution ray-sampling algorithm, we train the NeRF model using images with $\frac{1}{8}$ resolution for one epoch, images with $\frac{1}{4}$ resolution for one epoch, images with $\frac{1}{2}$ resolution for one epoch, and then original images for the rest. PyNeRF shows that this algorithm can accelerate training by using the model initialized in the coarse-grained training stage to guide the fine-grained training stage. However, objects from the Blender dataset have simple geometry, which makes it easy to initialize the NeRF model, leading to tiny acceleration.



\subsection{Results After Convergence}
\label{subsec:convergence}

    \begin{table}
    \caption{Results after convergence on the bicycle scene.}
    \label{tab:convergence}
    \centering
    \begin{tabular}{ccc}
        \toprule
        Dataset Split & Test & Train \\
        \cmidrule{1-3}
        Method & PSNR / SSIM / LPIPS & PSNR / SSIM / LPIPS \\
        \cmidrule{1-1}
        \cmidrule{2-2}
        \cmidrule{3-3}
        No Strategy & 28.01 / 0.853 / 0.081 & 30.02 / 0.917 / 0.081\\
        MCBlock & 27.84 / 0.845 / 0.082 & 30.94 / 0.923 / 0.052 \\
        \bottomrule
    \end{tabular}
    \end{table}

    In Table~\ref{tab:convergence} we provide the results after the testing PSNR converges (100000 iterations), on the bicycle scene from the Mipnerf360 dataset. On testing views, MCBlock achieves results slightly worse than the baselines, while at the same time achieving much better results on the training views. It's because the estimated block partition of the testing views is not that precise, leading to deterioration in testing views. Besides, our main contribution of acceleration focuses on the early stages of training, as shown in Table~\ref{tab:diff_algorithm}, instead of the performance after convergence.

\subsection{Ablation Study}
\label{subsec:ablation_study}

    \begin{figure*}
    \centering
        \begin{subfigure}[t]{0.3\textwidth}
            \centering
            \includegraphics[width=\linewidth]{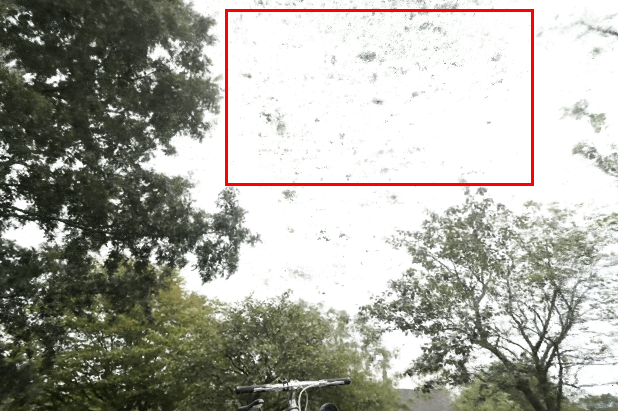}
            \caption{W/o Block Partition \& W/o Block Rendering}
            \label{wo_partition_wo_rendering}
        \end{subfigure}
        \hfill
        \begin{subfigure}[t]{0.3\textwidth}
            \centering
            \includegraphics[width=\linewidth]{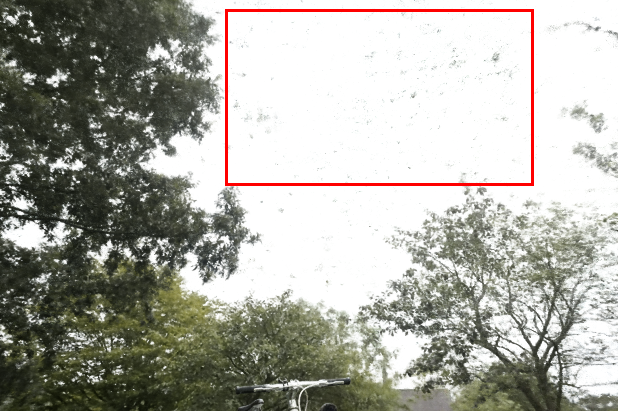}
            \caption{W/ Block Partition \& W/o Block Rendering}
            \label{w_partition_wo_rendering}
        \end{subfigure}
        \hfill
        \begin{subfigure}[t]{0.3\textwidth}
            \centering
            \includegraphics[width=\linewidth]{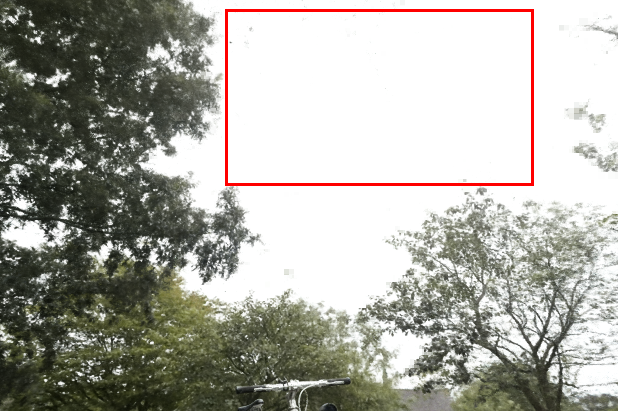}
            \caption{W/ Block Partition \& W/ Block Rendering}
            \label{w_partition_w_rendering}
        \end{subfigure}
    \caption{Rendered images with different strategy combinations. After applying the block partition strategy, the floaters are partially removed. After applying the block rendering strategy, the floaters are completely removed.}
    \label{ablation_figure}
    \end{figure*}

    \begin{table}
    \caption{LPIPS of each ablation at each iteration. Removing any module leads to a decrease in performance.}
    \label{ablation}
    \centering
    \resizebox{\linewidth}{!}
    {
        \begin{tabu}{*{6}{c}}
            \toprule
            Iteration & 1000 & 5000 & 9000 & 13000 & 17000 \\
            \cmidrule{1-6}
            Method & \multicolumn{5}{c}{LPIPS} \\
            \cmidrule(r){1-1}
            \cmidrule(r){2-6}
            MCBlock & \textbf{0.242} & \textbf{0.082} & \textbf{0.066} & \textbf{0.058} & \textbf{0.052} \\
            W/o Initialization & 0.249 & 0.096 & 0.071 & 0.061 & 0.056 \\
            W/o Block Partition & 0.243 & \textbf{0.082} & 0.069 & 0.062 & 0.056 \\
            W/o Block Selection & 0.245 & 0.093 & 0.070 & 0.060 & 0.054 \\
            W/o Block Rendering & 0.249 & 0.085 & 0.068 & \textbf{0.058} & 0.053 \\
            \bottomrule
        \end{tabu}
    }
    \end{table}




    
    We evaluate the utilities of four parts of our algorithm, namely the texture-based \textbf{\textit{tree structure initialization}}, the dynamic block pruning and expansion (\textbf{\textit{block partition}}), the UCT-guided block selection (\textbf{\textit{block selection}}), and the block-wise novel view synthesis (\textbf{\textit{block rendering}}). Results are shown in Table~\ref{ablation}.


    \textbf{Tree Structure Initialization Ablation}: Without the texture-based tree structure initialization, the MCTS-based training process starts from only root nodes. Although in the later training stage, its rendering quality gradually catches up, in the beginning, its LPIPS is high. It's because, at the start, training blocks too large for an image region can only optimize the extremely low-frequency parameters, which is not enough for rendering high-frequency details.
    
    
    \textbf{Block Partition Ablation}: Without the block partition strategy, we use only the color variance as the partition criterion, which leads to a static block partition. In this way, the high-frequency parameters in simple-texture regions are always unsupervised, leading to large amounts of high-frequency artifacts in rendered images. As illustrated in Figure~\ref{wo_partition_wo_rendering} and Figure~\ref{w_partition_wo_rendering}, our block partition strategy can eliminate some of the floaters in the sky region.
    
    
    \textbf{Block Selection Ablation}: Without the block selection strategy, we sample the blocks with the same probability. This still possesses a sampling bias towards complex-texture regions, because there are usually more blocks in complex-texture regions than in simple-texture regions. But among the complex-texture regions, it is close to the random sampling, which conducts unnecessary sampling in well-trained regions, leading to a decline in training efficiency.
    
    
    \textbf{Block Rendering Ablation}: Without the block rendering strategy, we conduct novel view synthesis through pixel-wise rendering like most NeRF works. It can be seen in Figure~\ref{w_partition_wo_rendering} that, although the block splitting can eliminate artifacts when rendering training images, the faulty parameters remaining in the model result in artifacts during novel view synthesis. But rendering simple-texture regions with large blocks only uses low-frequency model parameters, which are well-trained during training, leading to artifact-free rendered images, as shown in Figure~\ref{w_partition_w_rendering}.

\subsection{Time Consumption}
\label{subsec:time_conspt}

    \begin{figure}
        \centering
        \begin{subfigure}{0.45\linewidth}
            \centering
            \includegraphics[width=\linewidth]{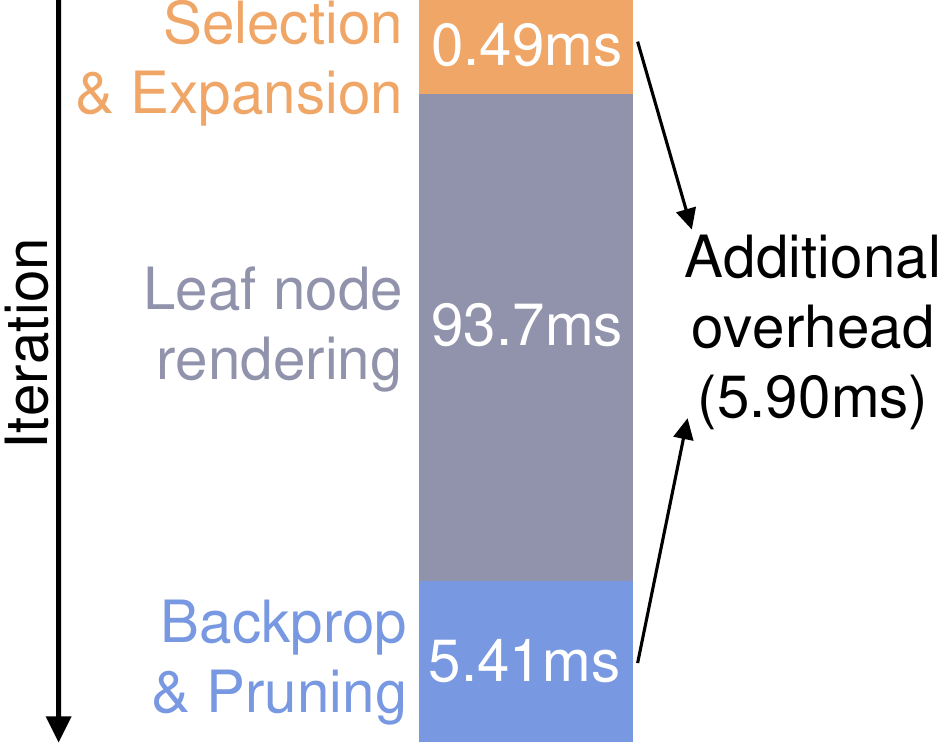}
            \caption{Zipnerf Overhead}
            \label{subfig:time_bar_zipnerf}
            \vspace{2mm}
        \end{subfigure}
        \hfill
        \begin{subfigure}{0.45\linewidth}
            \centering
            \includegraphics[width=\linewidth]{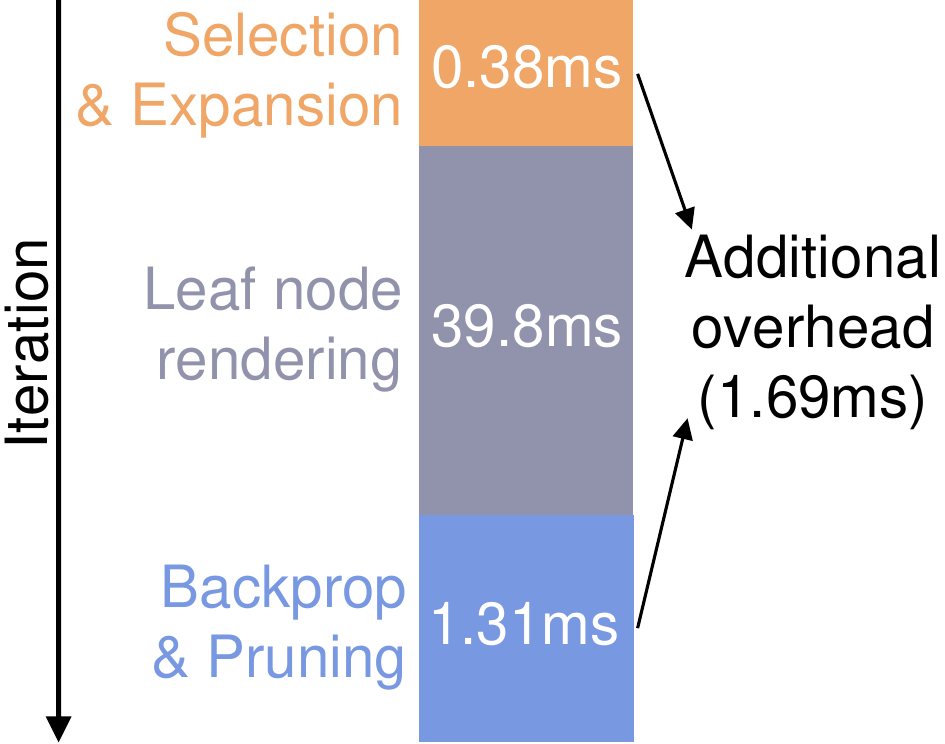}
            \caption{Mipnerf Overhead}
            \label{subfig:time_bar_mipnerf}
            \vspace{2mm}
        \end{subfigure}
        \begin{subfigure}{0.45\linewidth}
            \centering
            \includegraphics[width=\linewidth]{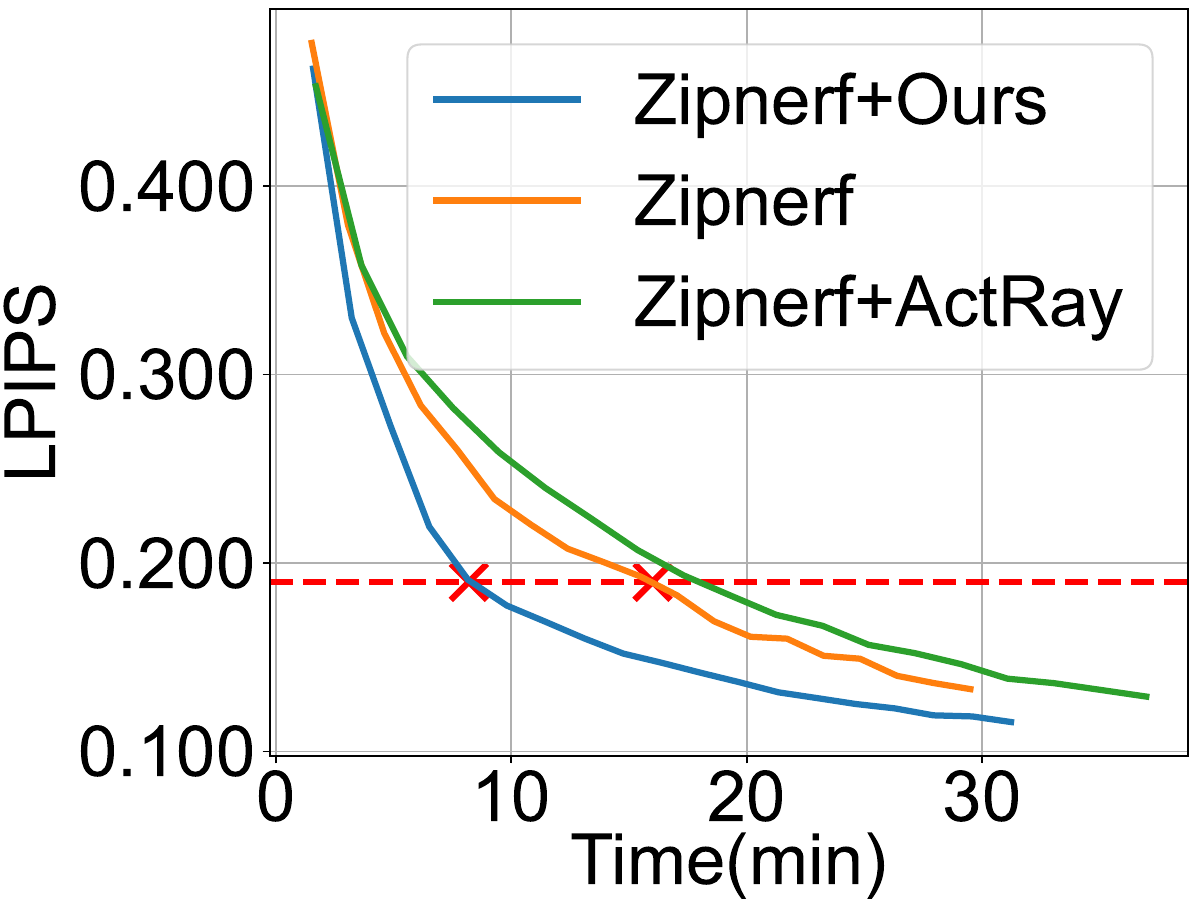}
            \caption{Zipnerf Acceleration}
            \label{subfig:time_lpips_zipnerf}
        \end{subfigure}
        \hfill
        \begin{subfigure}{0.45\linewidth}
            \centering
            \includegraphics[width=\linewidth]{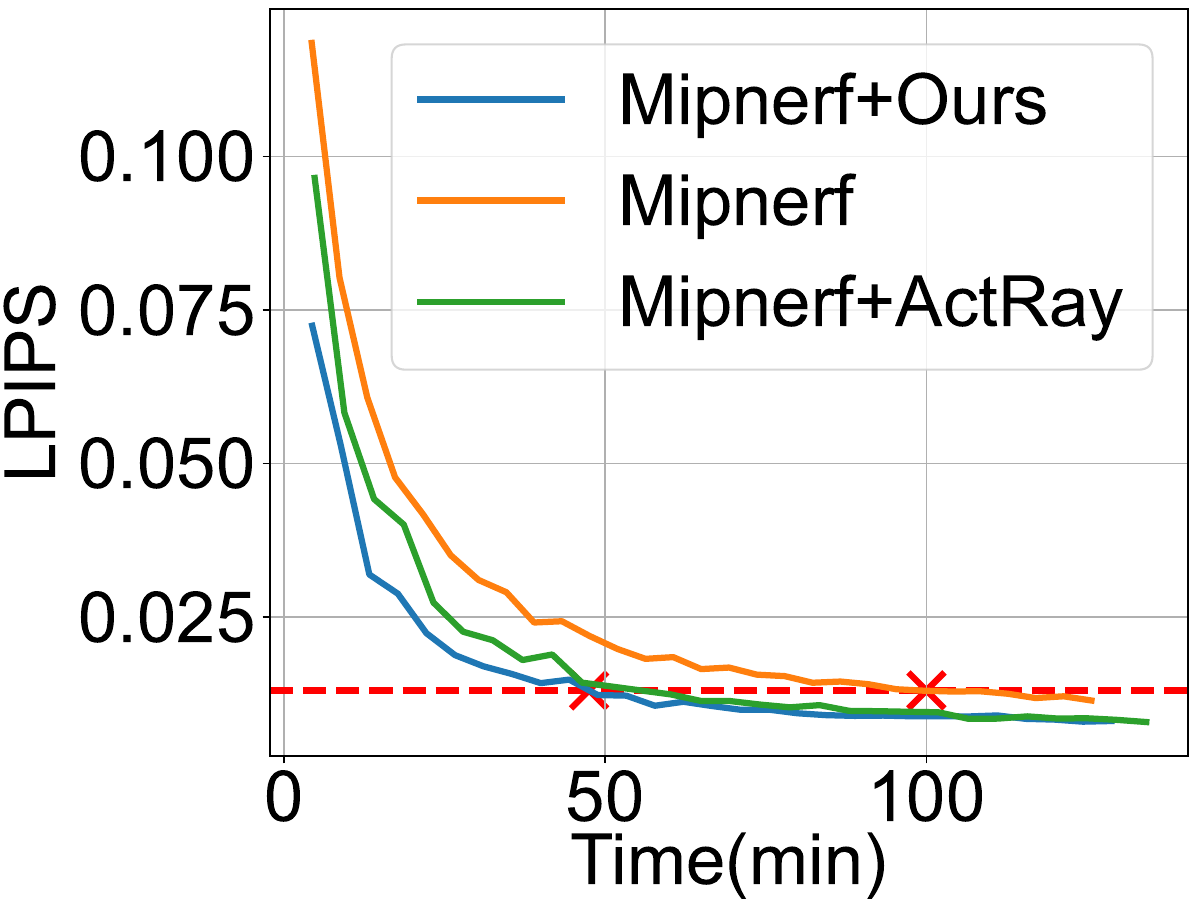}
            \caption{Mipenrf Acceleration}
            \label{subfig:time_lpips_mipnerf}
        \end{subfigure}
        \caption{We measure the time consumption of each procedure of one iteration. We include \textbf{(a)} only \textbf{5.9\%} for the Zipnerf model on the bicycle scene and \textbf{(b)} only \textbf{4.1\%} for the Mipnerf model on the lego scene. This promises our advantage over random selection and ActRay. \textbf{(c)} For the Zipnerf model on the bicycle scene, MCBlock can reduce the training time to \textbf{51.3\%} compared with the random ray selection strategy. \textbf{(d)} For the Mipnerf model on the lego scene, MCBlock can reduce the training time to \textbf{47.5\%}.}
        \label{fig:time}
    \end{figure}

    We measure the time consumption of MCBlock and ActRay implemented in Nerfstudio on a single NVIDIA RTX 4090. The batch size is 8192 for Zipnerf and 1024 for Mipnerf. Figure~\ref{subfig:time_bar_zipnerf} and Figure~\ref{subfig:time_bar_mipnerf} show that MCBlock only holds a tiny overhead, which only accounts for \textbf{5.9\%} of the total iteration time for Zipnerf model and \textbf{4.1\%} for Mipnerf model.
    
    This enables MCBlock to keep the training acceleration with the time as the horizontal axis. As shown in Figure~\ref{subfig:time_lpips_zipnerf} and Figure~\ref{subfig:time_lpips_mipnerf}, it can reduce the time to \textbf{47.5\%} while achieving the same quality. However, ActRay even decreases the training speed. There are two reasons: First, the loss propagation strategy of ActRay fails under outdoor scenes due to inaccurate depth estimation. Second, ActRay poses a large additional overhead of 24.8ms (for Zipnerf), which accounts for 20.9\% of the total iteration time.




\section{Conclusion}
\label{sec:conclusion}

    \textbf{Limitation}: The advantage of MCBlock over other ray-sampling algorithms on the training set is much more noticeable than in novel view synthesis. That's because the block-wise rendering strategy still slightly relies on the learned geometry. We may employ sparsity prior like SNeRG~\cite{Hedman_2021_ICCV} to eliminate more high-frequency flaws, and apply pixel-wise rendering instead of block-wise rendering.

    We propose an MCTS-based dynamic-resolution ray-sampling algorithm, MCBlock, to accelerate NeRF training. It combines active sampling and dynamic-resolution sampling to improve the reconstruction quality of complex-texture regions while removing high-frequency artifacts for simple-texture regions. The experiments show that MCBlock can accelerate NeRF training by up to \textbf{2.33x}, surpassing other ray-sampling algorithms. We hope MCBlock can bring us closer to real-time NeRF applications.

\bibliographystyle{ACM-Reference-Format}
\bibliography{sample-base}


\begin{thebibliography}{50}


\ifx \showCODEN    \undefined \def \showCODEN     #1{\unskip}     \fi
\ifx \showISBNx    \undefined \def \showISBNx     #1{\unskip}     \fi
\ifx \showISBNxiii \undefined \def \showISBNxiii  #1{\unskip}     \fi
\ifx \showISSN     \undefined \def \showISSN      #1{\unskip}     \fi
\ifx \showLCCN     \undefined \def \showLCCN      #1{\unskip}     \fi
\ifx \shownote     \undefined \def \shownote      #1{#1}          \fi
\ifx \showarticletitle \undefined \def \showarticletitle #1{#1}   \fi
\ifx \showURL      \undefined \def \showURL       {\relax}        \fi
\providecommand\bibfield[2]{#2}
\providecommand\bibinfo[2]{#2}
\providecommand\natexlab[1]{#1}
\providecommand\showeprint[2][]{arXiv:#2}

\bibitem[Bao et~al\mbox{.}(2023)]%
        {bao2023sine}
\bibfield{author}{\bibinfo{person}{Chong Bao}, \bibinfo{person}{Yinda Zhang}, \bibinfo{person}{Bangbang Yang}, \bibinfo{person}{Tianxing Fan}, \bibinfo{person}{Zesong Yang}, \bibinfo{person}{Hujun Bao}, \bibinfo{person}{Guofeng Zhang}, {and} \bibinfo{person}{Zhaopeng Cui}.} \bibinfo{year}{2023}\natexlab{}.
\newblock \showarticletitle{Sine: Semantic-driven image-based nerf editing with prior-guided editing field}. In \bibinfo{booktitle}{\emph{Proceedings of the IEEE/CVF Conference on Computer Vision and Pattern Recognition}}. \bibinfo{pages}{20919--20929}.
\newblock


\bibitem[Barron et~al\mbox{.}(2021)]%
        {Barron_2021_ICCV}
\bibfield{author}{\bibinfo{person}{Jonathan~T. Barron}, \bibinfo{person}{Ben Mildenhall}, \bibinfo{person}{Matthew Tancik}, \bibinfo{person}{Peter Hedman}, \bibinfo{person}{Ricardo Martin-Brualla}, {and} \bibinfo{person}{Pratul~P. Srinivasan}.} \bibinfo{year}{2021}\natexlab{}.
\newblock \showarticletitle{Mip-NeRF: A Multiscale Representation for Anti-Aliasing Neural Radiance Fields}. In \bibinfo{booktitle}{\emph{Proceedings of the IEEE/CVF International Conference on Computer Vision (ICCV)}}. \bibinfo{pages}{5855--5864}.
\newblock


\bibitem[Barron et~al\mbox{.}(2022)]%
        {Barron_2022_CVPR}
\bibfield{author}{\bibinfo{person}{Jonathan~T. Barron}, \bibinfo{person}{Ben Mildenhall}, \bibinfo{person}{Dor Verbin}, \bibinfo{person}{Pratul~P. Srinivasan}, {and} \bibinfo{person}{Peter Hedman}.} \bibinfo{year}{2022}\natexlab{}.
\newblock \showarticletitle{Mip-NeRF 360: Unbounded Anti-Aliased Neural Radiance Fields}. In \bibinfo{booktitle}{\emph{Proceedings of the IEEE/CVF Conference on Computer Vision and Pattern Recognition (CVPR)}}. \bibinfo{pages}{5470--5479}.
\newblock


\bibitem[Barron et~al\mbox{.}(2023)]%
        {Barron_2023_ICCV}
\bibfield{author}{\bibinfo{person}{Jonathan~T. Barron}, \bibinfo{person}{Ben Mildenhall}, \bibinfo{person}{Dor Verbin}, \bibinfo{person}{Pratul~P. Srinivasan}, {and} \bibinfo{person}{Peter Hedman}.} \bibinfo{year}{2023}\natexlab{}.
\newblock \showarticletitle{Zip-NeRF: Anti-Aliased Grid-Based Neural Radiance Fields}. In \bibinfo{booktitle}{\emph{Proceedings of the IEEE/CVF International Conference on Computer Vision (ICCV)}}. \bibinfo{pages}{19697--19705}.
\newblock


\bibitem[Bello et~al\mbox{.}(2024)]%
        {bello2024pronerf}
\bibfield{author}{\bibinfo{person}{Juan Luis~Gonzalez Bello}, \bibinfo{person}{Minh-Quan~Viet Bui}, {and} \bibinfo{person}{Munchurl Kim}.} \bibinfo{year}{2024}\natexlab{}.
\newblock \showarticletitle{ProNeRF: Learning Efficient Projection-Aware Ray Sampling for Fine-Grained Implicit Neural Radiance Fields}.
\newblock \bibinfo{journal}{\emph{IEEE Access}} (\bibinfo{year}{2024}).
\newblock


\bibitem[Boss et~al\mbox{.}(2020)]%
        {Boss20arxiv_NeRD}
\bibfield{author}{\bibinfo{person}{Mark Boss}, \bibinfo{person}{Raphael Braun}, \bibinfo{person}{Varun Jampani}, \bibinfo{person}{Jonathan~T. Barron}, \bibinfo{person}{Ce Liu}, {and} \bibinfo{person}{Hendrik Lensch}.} \bibinfo{year}{2020}\natexlab{}.
\newblock \showarticletitle{{NeRD}: Neural Reflectance Decomposition from Image Collections}.
\newblock \bibinfo{journal}{\emph{https://arxiv.org/abs/2012.03918}} (\bibinfo{year}{2020}).
\newblock


\bibitem[Brosse et~al\mbox{.}(2018)]%
        {brosse2018promises}
\bibfield{author}{\bibinfo{person}{Nicolas Brosse}, \bibinfo{person}{Alain Durmus}, {and} \bibinfo{person}{Eric Moulines}.} \bibinfo{year}{2018}\natexlab{}.
\newblock \showarticletitle{The promises and pitfalls of stochastic gradient Langevin dynamics}.
\newblock \bibinfo{journal}{\emph{Advances in Neural Information Processing Systems}}  \bibinfo{volume}{31} (\bibinfo{year}{2018}).
\newblock


\bibitem[Chen et~al\mbox{.}(2022)]%
        {chen2022tensorf}
\bibfield{author}{\bibinfo{person}{Anpei Chen}, \bibinfo{person}{Zexiang Xu}, \bibinfo{person}{Andreas Geiger}, \bibinfo{person}{Jingyi Yu}, {and} \bibinfo{person}{Hao Su}.} \bibinfo{year}{2022}\natexlab{}.
\newblock \showarticletitle{Tensorf: Tensorial radiance fields}. In \bibinfo{booktitle}{\emph{European Conference on Computer Vision}}. Springer, \bibinfo{pages}{333--350}.
\newblock


\bibitem[Coulom(2007)]%
        {10.1007/978-3-540-75538-8_7}
\bibfield{author}{\bibinfo{person}{R{\'e}mi Coulom}.} \bibinfo{year}{2007}\natexlab{}.
\newblock \showarticletitle{Efficient Selectivity and Backup Operators in Monte-Carlo Tree Search}. In \bibinfo{booktitle}{\emph{Computers and Games}}, \bibfield{editor}{\bibinfo{person}{H.~Jaap van~den Herik}, \bibinfo{person}{Paolo Ciancarini}, {and} \bibinfo{person}{H.~H. L. M.~(Jeroen) Donkers}} (Eds.). \bibinfo{publisher}{Springer Berlin Heidelberg}, \bibinfo{address}{Berlin, Heidelberg}, \bibinfo{pages}{72--83}.
\newblock
\showISBNx{978-3-540-75538-8}


\bibitem[Devernay and Faugeras(1994)]%
        {323831}
\bibfield{author}{\bibinfo{person}{Devernay} {and} \bibinfo{person}{Faugeras}.} \bibinfo{year}{1994}\natexlab{}.
\newblock \showarticletitle{Computing differential properties of 3-D shapes from stereoscopic images without 3-D models}. In \bibinfo{booktitle}{\emph{1994 Proceedings of IEEE Conference on Computer Vision and Pattern Recognition}}. \bibinfo{pages}{208--213}.
\newblock
\href{https://doi.org/10.1109/CVPR.1994.323831}{doi:\nolinkurl{10.1109/CVPR.1994.323831}}


\bibitem[Fridovich-Keil et~al\mbox{.}(2022)]%
        {fridovich2022plenoxels}
\bibfield{author}{\bibinfo{person}{Sara Fridovich-Keil}, \bibinfo{person}{Alex Yu}, \bibinfo{person}{Matthew Tancik}, \bibinfo{person}{Qinhong Chen}, \bibinfo{person}{Benjamin Recht}, {and} \bibinfo{person}{Angjoo Kanazawa}.} \bibinfo{year}{2022}\natexlab{}.
\newblock \showarticletitle{Plenoxels: Radiance fields without neural networks}. In \bibinfo{booktitle}{\emph{Proceedings of the IEEE/CVF Conference on Computer Vision and Pattern Recognition}}. \bibinfo{pages}{5501--5510}.
\newblock


\bibitem[Fukuda et~al\mbox{.}(2024)]%
        {fukuda2024important}
\bibfield{author}{\bibinfo{person}{Kohei Fukuda}, \bibinfo{person}{Takio Kurita}, {and} \bibinfo{person}{Hiroaki Aizawa}.} \bibinfo{year}{2024}\natexlab{}.
\newblock \showarticletitle{Important Pixels Sampling for NeRF Training Based on Edge Values and Squared Errors Between the Ground Truth and the Estimated Colors}.
\newblock \bibinfo{journal}{\emph{Proceedings Copyright}}  \bibinfo{volume}{102} (\bibinfo{year}{2024}), \bibinfo{pages}{111}.
\newblock


\bibitem[Haque et~al\mbox{.}(2023)]%
        {instructnerf2023}
\bibfield{author}{\bibinfo{person}{Ayaan Haque}, \bibinfo{person}{Matthew Tancik}, \bibinfo{person}{Alexei Efros}, \bibinfo{person}{Aleksander Holynski}, {and} \bibinfo{person}{Angjoo Kanazawa}.} \bibinfo{year}{2023}\natexlab{}.
\newblock \showarticletitle{Instruct-NeRF2NeRF: Editing 3D Scenes with Instructions}. In \bibinfo{booktitle}{\emph{Proceedings of the IEEE/CVF International Conference on Computer Vision}}.
\newblock


\bibitem[Hedman et~al\mbox{.}(2021)]%
        {Hedman_2021_ICCV}
\bibfield{author}{\bibinfo{person}{Peter Hedman}, \bibinfo{person}{Pratul~P. Srinivasan}, \bibinfo{person}{Ben Mildenhall}, \bibinfo{person}{Jonathan~T. Barron}, {and} \bibinfo{person}{Paul Debevec}.} \bibinfo{year}{2021}\natexlab{}.
\newblock \showarticletitle{Baking Neural Radiance Fields for Real-Time View Synthesis}. In \bibinfo{booktitle}{\emph{Proceedings of the IEEE/CVF International Conference on Computer Vision (ICCV)}}. \bibinfo{pages}{5875--5884}.
\newblock


\bibitem[Hu et~al\mbox{.}(2022)]%
        {hu2022efficientnerf}
\bibfield{author}{\bibinfo{person}{Tao Hu}, \bibinfo{person}{Shu Liu}, \bibinfo{person}{Yilun Chen}, \bibinfo{person}{Tiancheng Shen}, {and} \bibinfo{person}{Jiaya Jia}.} \bibinfo{year}{2022}\natexlab{}.
\newblock \showarticletitle{Efficientnerf efficient neural radiance fields}. In \bibinfo{booktitle}{\emph{Proceedings of the IEEE/CVF Conference on Computer Vision and Pattern Recognition}}. \bibinfo{pages}{12902--12911}.
\newblock


\bibitem[Huang et~al\mbox{.}(2023)]%
        {huang2023local}
\bibfield{author}{\bibinfo{person}{Xin Huang}, \bibinfo{person}{Qi Zhang}, \bibinfo{person}{Ying Feng}, \bibinfo{person}{Xiaoyu Li}, \bibinfo{person}{Xuan Wang}, {and} \bibinfo{person}{Qing Wang}.} \bibinfo{year}{2023}\natexlab{}.
\newblock \showarticletitle{Local implicit ray function for generalizable radiance field representation}. In \bibinfo{booktitle}{\emph{Proceedings of the IEEE/CVF Conference on Computer Vision and Pattern Recognition}}. \bibinfo{pages}{97--107}.
\newblock


\bibitem[Isaac-Medina et~al\mbox{.}(2023)]%
        {isaac2023exact}
\bibfield{author}{\bibinfo{person}{Brian~KS Isaac-Medina}, \bibinfo{person}{Chris~G Willcocks}, {and} \bibinfo{person}{Toby~P Breckon}.} \bibinfo{year}{2023}\natexlab{}.
\newblock \showarticletitle{Exact-nerf: An exploration of a precise volumetric parameterization for neural radiance fields}. In \bibinfo{booktitle}{\emph{Proceedings of the IEEE/CVF Conference on Computer Vision and Pattern Recognition}}. \bibinfo{pages}{66--75}.
\newblock


\bibitem[Jiakai et~al\mbox{.}(2021)]%
        {zhang2021stnerf}
\bibfield{author}{\bibinfo{person}{Zhang Jiakai}, \bibinfo{person}{Liu Xinhang}, \bibinfo{person}{Ye Xinyi}, \bibinfo{person}{Zhao Fuqiang}, \bibinfo{person}{Zhang Yanshun}, \bibinfo{person}{Wu Minye}, \bibinfo{person}{Zhang Yingliang}, \bibinfo{person}{Xu Lan}, {and} \bibinfo{person}{Yu Jingyi}.} \bibinfo{year}{2021}\natexlab{}.
\newblock \showarticletitle{Editable Free-Viewpoint Video using a Layered Neural Representation}. In \bibinfo{booktitle}{\emph{ACM SIGGRAPH}}.
\newblock


\bibitem[Jin et~al\mbox{.}(2023)]%
        {jin2023tensoir}
\bibfield{author}{\bibinfo{person}{Haian Jin}, \bibinfo{person}{Isabella Liu}, \bibinfo{person}{Peijia Xu}, \bibinfo{person}{Xiaoshuai Zhang}, \bibinfo{person}{Songfang Han}, \bibinfo{person}{Sai Bi}, \bibinfo{person}{Xiaowei Zhou}, \bibinfo{person}{Zexiang Xu}, {and} \bibinfo{person}{Hao Su}.} \bibinfo{year}{2023}\natexlab{}.
\newblock \showarticletitle{Tensoir: Tensorial inverse rendering}. In \bibinfo{booktitle}{\emph{Proceedings of the IEEE/CVF Conference on Computer Vision and Pattern Recognition}}. \bibinfo{pages}{165--174}.
\newblock


\bibitem[Kato and Tarashima(2024)]%
        {kato2024plug}
\bibfield{author}{\bibinfo{person}{Yoshio Kato} {and} \bibinfo{person}{Shuhei Tarashima}.} \bibinfo{year}{2024}\natexlab{}.
\newblock \showarticletitle{Plug-and-Play Acceleration of Occupancy Grid-based NeRF Rendering using VDB Grid and Hierarchical Ray Traversal}.
\newblock \bibinfo{journal}{\emph{arXiv preprint arXiv:2404.10272}} (\bibinfo{year}{2024}).
\newblock


\bibitem[Kerbl et~al\mbox{.}(2023)]%
        {kerbl20233d}
\bibfield{author}{\bibinfo{person}{Bernhard Kerbl}, \bibinfo{person}{Georgios Kopanas}, \bibinfo{person}{Thomas Leimk{\"u}hler}, {and} \bibinfo{person}{George Drettakis}.} \bibinfo{year}{2023}\natexlab{}.
\newblock \showarticletitle{3d gaussian splatting for real-time radiance field rendering}.
\newblock \bibinfo{journal}{\emph{ACM Transactions on Graphics}} \bibinfo{volume}{42}, \bibinfo{number}{4} (\bibinfo{year}{2023}), \bibinfo{pages}{1--14}.
\newblock


\bibitem[Kheradmand et~al\mbox{.}(2024)]%
        {kheradmand2024accelerating}
\bibfield{author}{\bibinfo{person}{Shakiba Kheradmand}, \bibinfo{person}{Daniel Rebain}, \bibinfo{person}{Gopal Sharma}, \bibinfo{person}{Hossam Isack}, \bibinfo{person}{Abhishek Kar}, \bibinfo{person}{Andrea Tagliasacchi}, {and} \bibinfo{person}{Kwang~Moo Yi}.} \bibinfo{year}{2024}\natexlab{}.
\newblock \showarticletitle{Accelerating Neural Field Training via Soft Mining}. In \bibinfo{booktitle}{\emph{Proceedings of the IEEE/CVF Conference on Computer Vision and Pattern Recognition}}. \bibinfo{pages}{20071--20080}.
\newblock


\bibitem[Kocsis and Szepesv{\'a}ri(2006)]%
        {10.1007/11871842_29}
\bibfield{author}{\bibinfo{person}{Levente Kocsis} {and} \bibinfo{person}{Csaba Szepesv{\'a}ri}.} \bibinfo{year}{2006}\natexlab{}.
\newblock \showarticletitle{Bandit Based Monte-Carlo Planning}. In \bibinfo{booktitle}{\emph{Machine Learning: ECML 2006}}, \bibfield{editor}{\bibinfo{person}{Johannes F{\"u}rnkranz}, \bibinfo{person}{Tobias Scheffer}, {and} \bibinfo{person}{Myra Spiliopoulou}} (Eds.). \bibinfo{publisher}{Springer Berlin Heidelberg}, \bibinfo{address}{Berlin, Heidelberg}, \bibinfo{pages}{282--293}.
\newblock
\showISBNx{978-3-540-46056-5}


\bibitem[Korhonen et~al\mbox{.}(2025)]%
        {korhonen2025efficient}
\bibfield{author}{\bibinfo{person}{Juuso Korhonen}, \bibinfo{person}{Goutham Rangu}, \bibinfo{person}{Hamed~R Tavakoli}, {and} \bibinfo{person}{Juho Kannala}.} \bibinfo{year}{2025}\natexlab{}.
\newblock \showarticletitle{Efficient NeRF Optimization-Not All Samples Remain Equally Hard}. In \bibinfo{booktitle}{\emph{European Conference on Computer Vision}}. Springer, \bibinfo{pages}{198--213}.
\newblock


\bibitem[Li and Zhang(2024)]%
        {li2024l0}
\bibfield{author}{\bibinfo{person}{Liangchen Li} {and} \bibinfo{person}{Juyong Zhang}.} \bibinfo{year}{2024}\natexlab{}.
\newblock \showarticletitle{L0-Sampler: An L0 Model Guided Volume Sampling for NeRF}. In \bibinfo{booktitle}{\emph{Proceedings of the IEEE/CVF Conference on Computer Vision and Pattern Recognition}}. \bibinfo{pages}{21390--21400}.
\newblock


\bibitem[Li et~al\mbox{.}(2024)]%
        {li2024know}
\bibfield{author}{\bibinfo{person}{Rui Li}, \bibinfo{person}{Tobias Fischer}, \bibinfo{person}{Mattia Segu}, \bibinfo{person}{Marc Pollefeys}, \bibinfo{person}{Luc Van~Gool}, {and} \bibinfo{person}{Federico Tombari}.} \bibinfo{year}{2024}\natexlab{}.
\newblock \showarticletitle{Know Your Neighbors: Improving Single-View Reconstruction via Spatial Vision-Language Reasoning}. In \bibinfo{booktitle}{\emph{CVPR}}.
\newblock


\bibitem[Liu et~al\mbox{.}(2020a)]%
        {Liu2020Watch}
\bibfield{author}{\bibinfo{person}{Anji Liu}, \bibinfo{person}{Jianshu Chen}, \bibinfo{person}{Mingze Yu}, \bibinfo{person}{Yu Zhai}, \bibinfo{person}{Xuewen Zhou}, {and} \bibinfo{person}{Ji Liu}.} \bibinfo{year}{2020}\natexlab{a}.
\newblock \showarticletitle{Watch the Unobserved: A Simple Approach to Parallelizing Monte Carlo Tree Search}. In \bibinfo{booktitle}{\emph{International Conference on Learning Representations}}.
\newblock
\urldef\tempurl%
\url{https://openreview.net/forum?id=BJlQtJSKDB}
\showURL{%
\tempurl}


\bibitem[Liu et~al\mbox{.}(2020b)]%
        {liu2020neural}
\bibfield{author}{\bibinfo{person}{Lingjie Liu}, \bibinfo{person}{Jiatao Gu}, \bibinfo{person}{Kyaw Zaw~Lin}, \bibinfo{person}{Tat-Seng Chua}, {and} \bibinfo{person}{Christian Theobalt}.} \bibinfo{year}{2020}\natexlab{b}.
\newblock \showarticletitle{Neural sparse voxel fields}.
\newblock \bibinfo{journal}{\emph{Advances in Neural Information Processing Systems}}  \bibinfo{volume}{33} (\bibinfo{year}{2020}), \bibinfo{pages}{15651--15663}.
\newblock


\bibitem[Mildenhall et~al\mbox{.}(2021)]%
        {10.1145/3503250}
\bibfield{author}{\bibinfo{person}{Ben Mildenhall}, \bibinfo{person}{Pratul~P. Srinivasan}, \bibinfo{person}{Matthew Tancik}, \bibinfo{person}{Jonathan~T. Barron}, \bibinfo{person}{Ravi Ramamoorthi}, {and} \bibinfo{person}{Ren Ng}.} \bibinfo{year}{2021}\natexlab{}.
\newblock \showarticletitle{NeRF: representing scenes as neural radiance fields for view synthesis}.
\newblock \bibinfo{journal}{\emph{Commun. ACM}} \bibinfo{volume}{65}, \bibinfo{number}{1} (\bibinfo{date}{dec} \bibinfo{year}{2021}), \bibinfo{pages}{99–106}.
\newblock
\showISSN{0001-0782}
\href{https://doi.org/10.1145/3503250}{doi:\nolinkurl{10.1145/3503250}}


\bibitem[Mouragnon et~al\mbox{.}(2006)]%
        {1640781}
\bibfield{author}{\bibinfo{person}{E. Mouragnon}, \bibinfo{person}{M. Lhuillier}, \bibinfo{person}{M. Dhome}, \bibinfo{person}{F. Dekeyser}, {and} \bibinfo{person}{P. Sayd}.} \bibinfo{year}{2006}\natexlab{}.
\newblock \showarticletitle{Real Time Localization and 3D Reconstruction}. In \bibinfo{booktitle}{\emph{2006 IEEE Computer Society Conference on Computer Vision and Pattern Recognition (CVPR'06)}}, Vol.~\bibinfo{volume}{1}. \bibinfo{pages}{363--370}.
\newblock
\href{https://doi.org/10.1109/CVPR.2006.236}{doi:\nolinkurl{10.1109/CVPR.2006.236}}


\bibitem[M{\"u}ller et~al\mbox{.}(2022)]%
        {muller2022instant}
\bibfield{author}{\bibinfo{person}{Thomas M{\"u}ller}, \bibinfo{person}{Alex Evans}, \bibinfo{person}{Christoph Schied}, {and} \bibinfo{person}{Alexander Keller}.} \bibinfo{year}{2022}\natexlab{}.
\newblock \showarticletitle{Instant neural graphics primitives with a multiresolution hash encoding}.
\newblock \bibinfo{journal}{\emph{ACM transactions on graphics (TOG)}} \bibinfo{volume}{41}, \bibinfo{number}{4} (\bibinfo{year}{2022}), \bibinfo{pages}{1--15}.
\newblock


\bibitem[M{\"u}ller et~al\mbox{.}(2021)]%
        {muller2021real}
\bibfield{author}{\bibinfo{person}{Thomas M{\"u}ller}, \bibinfo{person}{Fabrice Rousselle}, \bibinfo{person}{Jan Nov{\'a}k}, {and} \bibinfo{person}{Alexander Keller}.} \bibinfo{year}{2021}\natexlab{}.
\newblock \showarticletitle{Real-time neural radiance caching for path tracing}.
\newblock \bibinfo{journal}{\emph{arXiv preprint arXiv:2106.12372}} (\bibinfo{year}{2021}).
\newblock


\bibitem[Na et~al\mbox{.}(2024)]%
        {na2024uforecon}
\bibfield{author}{\bibinfo{person}{Youngju Na}, \bibinfo{person}{Woo~Jae Kim}, \bibinfo{person}{Kyu~Beom Han}, \bibinfo{person}{Suhyeon Ha}, {and} \bibinfo{person}{Sung-eui Yoon}.} \bibinfo{year}{2024}\natexlab{}.
\newblock \showarticletitle{UFORecon: Generalizable Sparse-View Surface Reconstruction from Arbitrary and UnFavOrable Data Sets}.
\newblock \bibinfo{journal}{\emph{arXiv preprint arXiv:2403.05086}} (\bibinfo{year}{2024}).
\newblock


\bibitem[Neal et~al\mbox{.}(2011)]%
        {neal2011mcmc}
\bibfield{author}{\bibinfo{person}{Radford~M Neal} {et~al\mbox{.}}} \bibinfo{year}{2011}\natexlab{}.
\newblock \showarticletitle{MCMC using Hamiltonian dynamics}.
\newblock \bibinfo{journal}{\emph{Handbook of markov chain monte carlo}} \bibinfo{volume}{2}, \bibinfo{number}{11} (\bibinfo{year}{2011}), \bibinfo{pages}{2}.
\newblock


\bibitem[Schwarz et~al\mbox{.}(2020)]%
        {Schwarz20neurips_graf}
\bibfield{author}{\bibinfo{person}{Katja Schwarz}, \bibinfo{person}{Yiyi Liao}, \bibinfo{person}{Michael Niemeyer}, {and} \bibinfo{person}{Andreas Geiger}.} \bibinfo{year}{2020}\natexlab{}.
\newblock \showarticletitle{Graf: Generative radiance fields for {3D}-aware image synthesis}. In \bibinfo{booktitle}{\emph{Advances in Neural Information Processing Systems (NeurIPS)}}, Vol.~\bibinfo{volume}{33}.
\newblock


\bibitem[Slivkins et~al\mbox{.}(2019)]%
        {slivkins2019introduction}
\bibfield{author}{\bibinfo{person}{Aleksandrs Slivkins} {et~al\mbox{.}}} \bibinfo{year}{2019}\natexlab{}.
\newblock \showarticletitle{Introduction to multi-armed bandits}.
\newblock \bibinfo{journal}{\emph{Foundations and Trends{\textregistered} in Machine Learning}} \bibinfo{volume}{12}, \bibinfo{number}{1-2} (\bibinfo{year}{2019}), \bibinfo{pages}{1--286}.
\newblock


\bibitem[Srinivasan et~al\mbox{.}(2020)]%
        {Srinivasan20arxiv_NeRV}
\bibfield{author}{\bibinfo{person}{Pratul Srinivasan}, \bibinfo{person}{Boyang Deng}, \bibinfo{person}{Xiuming Zhang}, \bibinfo{person}{Matthew Tancik}, \bibinfo{person}{Ben Mildenhall}, {and} \bibinfo{person}{Jonathan~T. Barron}.} \bibinfo{year}{2020}\natexlab{}.
\newblock \showarticletitle{{NeRV}: Neural Reflectance and Visibility Fields for Relighting and View Synthesis}.
\newblock \bibinfo{journal}{\emph{https://arxiv.org/abs/2012.03927}} (\bibinfo{year}{2020}).
\newblock


\bibitem[Sun et~al\mbox{.}(2022)]%
        {SunSC22}
\bibfield{author}{\bibinfo{person}{Cheng Sun}, \bibinfo{person}{Min Sun}, {and} \bibinfo{person}{Hwann{-}Tzong Chen}.} \bibinfo{year}{2022}\natexlab{}.
\newblock \showarticletitle{Direct Voxel Grid Optimization: Super-fast Convergence for Radiance Fields Reconstruction}. In \bibinfo{booktitle}{\emph{CVPR}}.
\newblock


\bibitem[Sun et~al\mbox{.}(2024)]%
        {sun2024efficient}
\bibfield{author}{\bibinfo{person}{Shilei Sun}, \bibinfo{person}{Ming Liu}, \bibinfo{person}{Zhongyi Fan}, \bibinfo{person}{Qingliang Jiao}, \bibinfo{person}{Yuxue Liu}, \bibinfo{person}{Liquan Dong}, {and} \bibinfo{person}{Lingqin Kong}.} \bibinfo{year}{2024}\natexlab{}.
\newblock \showarticletitle{Efficient ray sampling for radiance fields reconstruction}.
\newblock \bibinfo{journal}{\emph{Computers \& Graphics}}  \bibinfo{volume}{118} (\bibinfo{year}{2024}), \bibinfo{pages}{48--59}.
\newblock


\bibitem[Tancik et~al\mbox{.}(2023)]%
        {nerfstudio}
\bibfield{author}{\bibinfo{person}{Matthew Tancik}, \bibinfo{person}{Ethan Weber}, \bibinfo{person}{Evonne Ng}, \bibinfo{person}{Ruilong Li}, \bibinfo{person}{Brent Yi}, \bibinfo{person}{Justin Kerr}, \bibinfo{person}{Terrance Wang}, \bibinfo{person}{Alexander Kristoffersen}, \bibinfo{person}{Jake Austin}, \bibinfo{person}{Kamyar Salahi}, \bibinfo{person}{Abhik Ahuja}, \bibinfo{person}{David McAllister}, {and} \bibinfo{person}{Angjoo Kanazawa}.} \bibinfo{year}{2023}\natexlab{}.
\newblock \showarticletitle{Nerfstudio: A Modular Framework for Neural Radiance Field Development}. In \bibinfo{booktitle}{\emph{ACM SIGGRAPH 2023 Conference Proceedings}} \emph{(\bibinfo{series}{SIGGRAPH '23})}.
\newblock


\bibitem[Turki et~al\mbox{.}(2024)]%
        {turki2024pynerf}
\bibfield{author}{\bibinfo{person}{Haithem Turki}, \bibinfo{person}{Michael Zollh{\"o}fer}, \bibinfo{person}{Christian Richardt}, {and} \bibinfo{person}{Deva Ramanan}.} \bibinfo{year}{2024}\natexlab{}.
\newblock \showarticletitle{Pynerf: Pyramidal neural radiance fields}.
\newblock \bibinfo{journal}{\emph{Advances in Neural Information Processing Systems}}  \bibinfo{volume}{36} (\bibinfo{year}{2024}).
\newblock


\bibitem[Verbin et~al\mbox{.}(2022)]%
        {verbin2022ref}
\bibfield{author}{\bibinfo{person}{Dor Verbin}, \bibinfo{person}{Peter Hedman}, \bibinfo{person}{Ben Mildenhall}, \bibinfo{person}{Todd Zickler}, \bibinfo{person}{Jonathan~T Barron}, {and} \bibinfo{person}{Pratul~P Srinivasan}.} \bibinfo{year}{2022}\natexlab{}.
\newblock \showarticletitle{Ref-nerf: Structured view-dependent appearance for neural radiance fields}. In \bibinfo{booktitle}{\emph{2022 IEEE/CVF Conference on Computer Vision and Pattern Recognition (CVPR)}}. IEEE, \bibinfo{pages}{5481--5490}.
\newblock


\bibitem[Wang et~al\mbox{.}(2022)]%
        {wang2022clip}
\bibfield{author}{\bibinfo{person}{Can Wang}, \bibinfo{person}{Menglei Chai}, \bibinfo{person}{Mingming He}, \bibinfo{person}{Dongdong Chen}, {and} \bibinfo{person}{Jing Liao}.} \bibinfo{year}{2022}\natexlab{}.
\newblock \showarticletitle{Clip-nerf: Text-and-image driven manipulation of neural radiance fields}. In \bibinfo{booktitle}{\emph{Proceedings of the IEEE/CVF Conference on Computer Vision and Pattern Recognition}}. \bibinfo{pages}{3835--3844}.
\newblock


\bibitem[Wu et~al\mbox{.}(2023)]%
        {10.1145/3610548.3618254}
\bibfield{author}{\bibinfo{person}{Jiangkai Wu}, \bibinfo{person}{Liming Liu}, \bibinfo{person}{Yunpeng Tan}, \bibinfo{person}{Quanlu Jia}, \bibinfo{person}{Haodan Zhang}, {and} \bibinfo{person}{Xinggong Zhang}.} \bibinfo{year}{2023}\natexlab{}.
\newblock \showarticletitle{ActRay: Online Active Ray Sampling for Radiance Fields}. In \bibinfo{booktitle}{\emph{SIGGRAPH Asia 2023 Conference Papers}} (Sydney, NSW, Australia) \emph{(\bibinfo{series}{SA '23})}. Article \bibinfo{articleno}{97}, \bibinfo{numpages}{10}~pages.
\newblock


\bibitem[Wu et~al\mbox{.}(2024)]%
        {wu2024hi}
\bibfield{author}{\bibinfo{person}{Lizhou Wu}, \bibinfo{person}{Haozhe Zhu}, \bibinfo{person}{Jiapei Zheng}, \bibinfo{person}{Mengjie Li}, \bibinfo{person}{Yinuo Cheng}, \bibinfo{person}{Qi Liu}, \bibinfo{person}{Xiaoyang Zeng}, {and} \bibinfo{person}{Chixiao Chen}.} \bibinfo{year}{2024}\natexlab{}.
\newblock \showarticletitle{Hi-NeRF: A Multicore NeRF Accelerator With Hierarchical Empty Space Skipping for Edge 3-D Rendering}.
\newblock \bibinfo{journal}{\emph{IEEE Transactions on Very Large Scale Integration (VLSI) Systems}} (\bibinfo{year}{2024}).
\newblock


\bibitem[Yang et~al\mbox{.}(2024)]%
        {yang2024clear}
\bibfield{author}{\bibinfo{person}{WeiChen Yang}, \bibinfo{person}{JinLong Shi}, \bibinfo{person}{SuQin Bai}, \bibinfo{person}{Qiang Qian}, \bibinfo{person}{Zhen Ou}, \bibinfo{person}{Dan Xu}, \bibinfo{person}{Xin Shu}, {and} \bibinfo{person}{YunHan Sun}.} \bibinfo{year}{2024}\natexlab{}.
\newblock \showarticletitle{Clear-Plenoxels: Floaters free radiance fields without neural networks}.
\newblock \bibinfo{journal}{\emph{Knowledge-Based Systems}}  \bibinfo{volume}{299} (\bibinfo{year}{2024}), \bibinfo{pages}{112096}.
\newblock


\bibitem[Yong et~al\mbox{.}(2024)]%
        {yong2024gl}
\bibfield{author}{\bibinfo{person}{Silong Yong}, \bibinfo{person}{Yaqi Xie}, \bibinfo{person}{Simon Stepputtis}, {and} \bibinfo{person}{Katia Sycara}.} \bibinfo{year}{2024}\natexlab{}.
\newblock \showarticletitle{GL-NeRF: Gauss-Laguerre Quadrature Enables Training-Free NeRF Acceleration}.
\newblock \bibinfo{journal}{\emph{arXiv preprint arXiv:2410.19831}} (\bibinfo{year}{2024}).
\newblock


\bibitem[Yoo et~al\mbox{.}(2024)]%
        {yoo2024improving}
\bibfield{author}{\bibinfo{person}{Hye~Bin Yoo}, \bibinfo{person}{Hyun~Min Han}, \bibinfo{person}{Sung~Soo Hwang}, {and} \bibinfo{person}{Il~Yong Chun}.} \bibinfo{year}{2024}\natexlab{}.
\newblock \showarticletitle{Improving Neural Radiance Fields Using Near-Surface Sampling with Point Cloud Generation}.
\newblock \bibinfo{journal}{\emph{Neural Processing Letters}} \bibinfo{volume}{56}, \bibinfo{number}{4} (\bibinfo{year}{2024}), \bibinfo{pages}{214}.
\newblock


\bibitem[Yu et~al\mbox{.}(2020)]%
        {Yu20arxiv_pixelNeRF}
\bibfield{author}{\bibinfo{person}{Alex Yu}, \bibinfo{person}{Vickie Ye}, \bibinfo{person}{Matthew Tancik}, {and} \bibinfo{person}{Angjoo Kanazawa}.} \bibinfo{year}{2020}\natexlab{}.
\newblock \showarticletitle{{pixelNeRF}: Neural Radiance Fields from One or Few Images}.
\newblock \bibinfo{journal}{\emph{https://arxiv.org/abs/2012.02190}} (\bibinfo{year}{2020}).
\newblock


\bibitem[Zhang et~al\mbox{.}(2021)]%
        {zhang2021nerfactor}
\bibfield{author}{\bibinfo{person}{Xiuming Zhang}, \bibinfo{person}{Pratul~P Srinivasan}, \bibinfo{person}{Boyang Deng}, \bibinfo{person}{Paul Debevec}, \bibinfo{person}{William~T Freeman}, {and} \bibinfo{person}{Jonathan~T Barron}.} \bibinfo{year}{2021}\natexlab{}.
\newblock \showarticletitle{NeRFactor: Neural Factorization of Shape and Reflectance Under an Unknown Illumination}.
\newblock \bibinfo{journal}{\emph{https://arxiv.org/abs/2106.01970}} (\bibinfo{year}{2021}).
\newblock


\end{thebibliography}

\clearpage
\appendix
\section{Additional subjective results}
\label{sec:subjective_img}

\begin{figure}
\twocolumn[{
\renewcommand\twocolumn[1][]{#1}
  \centering
  \includegraphics[width=0.9\linewidth]{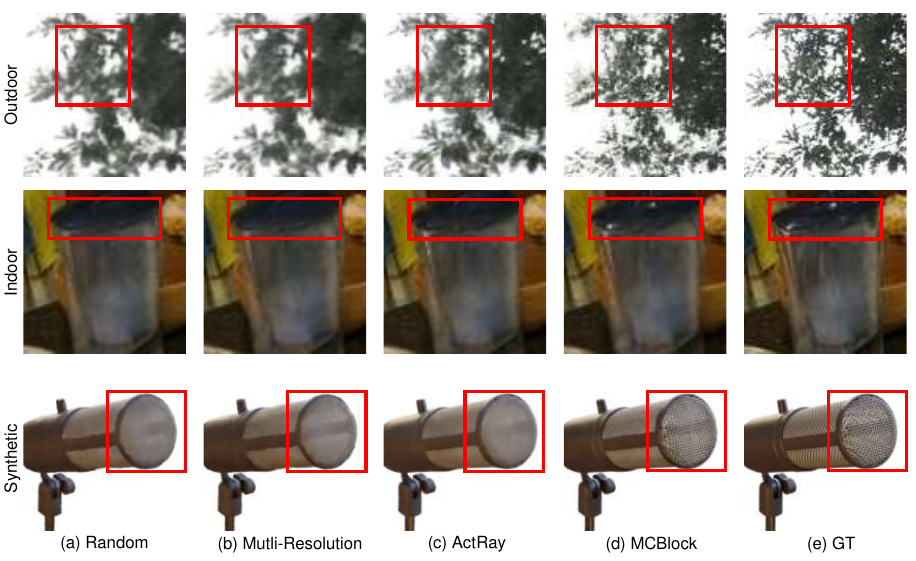}
  \caption{Rendered images with various ray-sampling algorithms: (a) random ray sampling in NeRF~\cite{10.1145/3503250}, (b) coarse-to-fine multi-resolution ray sampling~\cite{turki2024pynerf}. (c) active ray sampling (ActRay)~\cite{10.1145/3610548.3618254}, (d) our dynamic-resolution ray sampling (MCBlock) and (e) the ground truth. Rows respectively represent one outdoor scene (Outdoor), one indoor scene (Indoor) from the Mipnerf360 dataset~\cite{Barron_2022_CVPR}, and one synthetic scene (Synthetic) from the Blender dataset\cite{10.1145/3503250}. The proposed dynamic-resolution ray-sampling method MCBlock can accelerate the training speed up to 2.33x and render images with finer geometry and illumination details under the same training time.}
  \label{fig:teaser}
}]
\end{figure}

Subjective results can be found in Figure~\ref{fig:teaser}. From top to bottom, we show results of three different types of scenes, respectively the "bicycle" scene (an outdoor scene), the "counter" scene (an indoor scene), and the "mic" scene (a synthetic scene). From left to right, we show rendered results of the random ray-sampling algorithm, the coarse-to-fine multi-resolution ray-sampling algorithm of PyNeRF, the active ray-sampling algorithm of ActRay, and the active dynamic-resolution ray-sampling algorithm of our MCBlock, and also ground truth images. In the first line, the leaves gradually become clearer. In the second line, MCBlock provides finer illumination effects, such as the highlight of the bottle cap. In the third line, MCBlock renders the clearest texture of the microphone cover.

\end{document}